\documentclass[journal ]{new-aiaa}
\usepackage[utf8]{inputenc}
\usepackage{textcomp}

\usepackage{bm}
\usepackage{graphicx}
\usepackage{subcaption}
\usepackage{multirow}
\usepackage{amsmath}
\usepackage[version=4]{mhchem}
\usepackage{siunitx}
\usepackage{longtable,tabularx}
\title{Estimating Flow Velocity and Vehicle Angle-of-Attack from Non-invasive Piezoelectric Structural Measurements Using Deep Learning }

\author{Chandler B. Smith \footnote{Senior Member of Technical Staff, Simulation Modeling Sciences, Sandia National Laboratories}, S. Hales Swift, Andrew Steyer and Ihab El-Kady }
\affil{Sandia National Laboratories}

\begin{document}

\maketitle

\begin{abstract}
Accurate estimation of aerodynamic state variables such as freestream velocity and angle of attack (AoA) is important for aerodynamic load prediction, flight control, and model validation. This work presents a non-intrusive method for estimating vehicle velocity and AoA from structural vibration measurements rather than direct flow instrumentation such as pitot tubes. A dense array of piezoelectric sensors mounted on the interior skin of an aeroshell capture vibrations induced by turbulent boundary layer pressure fluctuations, and a convolutional neural network (CNN) is trained to invert these structural responses to recover velocity and AoA.

Proof-of-concept is demonstrated through controlled experiments in Sandia's hypersonic wind tunnel spanning zero and nonzero AoA configurations, Mach~5 and Mach~8 conditions, and both constant and continuously varying tunnel operations. The CNN is trained and evaluated using data from 16 wind tunnel runs, with a temporally centered held-out interval within each run used to form training, validation, and test datasets and assess intra-run temporal generalization. Raw CNN predictions exhibit increased variance during continuously varying conditions; a short-window moving-median post-processing step suppresses this variance and improves robustness. After post-processing, the method achieves a mean velocity error relative to the low-pass filtered reference velocity below 2.27~m/s (0.21\%) and a mean AoA error of $0.44^{\circ} (8.25\%)$ on held-out test data from the same experimental campaign, demonstrating feasibility of vibration-based velocity and AoA estimation in a controlled laboratory environment.
\end{abstract}

\section{Introduction}

Accurate estimates of aerodynamic state variables such as freestream velocity and angle of attack (AoA) are central to aerodynamic load prediction, flight control, and model validation. Conventional air data approaches (e.g., probes, vanes, or pressure port systems) can be difficult to integrate and calibrate on certain vehicle geometries and may disturb the local flow they seek to measure. These considerations motivate complementary sensing approaches that infer flow state indirectly from measurements taken within the vehicle structure. 

This paper investigates an internal sensing approach for estimating vehicle velocity and AoA using piezoelectric-based vibration measurements. The central premise is that the vibration signature of an aeroshell contains flow-dependent information through the coupling between turbulent boundary layer loading and structural dynamics. If successful, this approach could augment conventional air data instrumentation by providing additional, internal measurements that correlate with external flow state.

Indirect flow sensing is motivated by decades of experimental and theoretical work showing that turbulent boundary layers generate stochastic surface-pressure fields that excite structural vibration through fluctuating normal stresses at the fluid-solid interface \cite{Corcos1963,Willmarth1975,Bull1996,Chase1980,Bernardini2011,Smith2016,Casper2019}. Building on this coupling, several non-intrusive approaches have used structural vibration measurements to infer flow parameters in settings where direct flow field sensing is undesirable or impractical. For example, pipe vibration-based flow measurements have demonstrated correlations between wall acceleration and flow rate via (i) direct flow-induced vibrations \cite{Evans2004}, and (ii) flow-induced perturbations to mechanically excited plate waves \cite{Kim1996}. Thompson et al.\ \cite{Thompson2010} characterized pipe wall vibration as a function of flow speed and pipe geometry for fully developed turbulent flow, establishing empirical relationships between flow parameters and structural response. More recently, machine learning models have been used to extract features from vibration signals that correlate with flow parameters (e.g., mass-flow rate) that may be difficult to measure directly \cite{Kim2020}.

In the aerospace sector, fiber Bragg grating (FBG) sensors have emerged as a particularly promising technology for indirect flow parameter estimation through structural response measurements, offering a dense field of structural strain measurements and minimal weight additions to the structure of interest.
Nicolas et al. \cite{Nicolas2016} used a dense network of FBG sensors distributed along a carbon-composite wing to determine both in-flight loads and deflected wing shapes during flight operations. Wada et al. \cite{Wada2019} demonstrated a neural network approach wherein FBG strain data combined with the wing flap angles were used to estimate distributed wing loads and subsequently determine angle of attack, again without direct flow measurements. Furthermore, Kwon et al. \cite{Kwon2019} implemented an embedded FBG sensor network in a composite aircraft that successfully estimated wing loads during flight maneuvers. 
Collectively, these studies establish that internal strain/vibration measurements, when combined with data-driven inversion, can provide non-intrusive aerodynamic inference capabilities.

While FBG-based approaches have demonstrated effectiveness for load and AoA inference, the present work explores an alternative sensing technique using piezoelectric sensors to estimate freestream velocity and AoA. Piezoelectric sensors are another type of low-weight, non-intrusive strain measurement that generate voltage signals when mechanically deformed due to the material's inherent electromechanically coupled behavior. 
Our proposed method deploys a dense linear array of piezoelectric sensors, mounted to the interior skin of the aeroshell, to monitor vibration induced by turbulent boundary layer pressure fluctuations and thereby deduce the vehicle's AoA and velocity.  Because these interior-mounted sensors do not have direct access to the external flow, estimating freestream velocity and AoA becomes a challenging inverse problem.

We address this inverse problem using a convolutional neural network (CNN) trained on labeled wind tunnel data. Specifically, this work establishes proof-of-concept through a three-phase methodology: (i) acquire piezoelectric response data under controlled wind tunnel conditions where velocity and AoA are measured using conventional instrumentation, (ii) train a CNN to map windowed piezoelectric vibration measurements to velocity and AoA labels using supervised learning, and (iii) evaluate the trained model on withheld test data to assess intra-run temporal generalization capability. Raw network predictions are optionally refined using a short-window sliding median filter to suppress high-frequency output fluctuations and improve robustness.

Hypersonic wind tunnel experiments were used to exercise the methodology. We acquired tunnel freestream velocity estimates, AoA, and piezoelectric response data across sixteen runs spanning Mach~5 and Mach~8 conditions, multiple Reynolds-number states, and both constant and continuously varying tunnel operations. Piezoelectric signals were segmented into 16~ms windows and provided as input to a two-dimensional CNN that outputs estimates of velocity and AoA (via the inferred body-referenced velocity components). Performance is evaluated across the full set of runs, delineating between steady-state and time-varying tunnel conditions, to identify regimes where estimation is most reliable.

The remainder of this paper is organized as follows. Section \ref{sec:method_overview} details the proposed methodology and the neural network architecture. Section \ref{sec:hwt_data} describes the wind tunnel test setup, data acquisition, preprocessing procedures, and the range of flow conditions tested. Section \ref{sec:validation} presents the zero AoA velocity estimation results (Section \ref{sec:zeroaoa_free}), then extends the analysis to the combined AoA and velocity estimation problem at nonzero AoA (Section~\ref{sec:aoa}). Finally, Sections \ref{sec:summary} and \ref{sec:conclusion} summarizes our findings and outlines directions for future work.

\section{Method Overview}\label{sec:method_overview}

This work estimates aerodynamic quantities from internally mounted piezoelectric vibration measurements using a supervised CNN. A dense array of piezoelectric sensors mounted to the interior of a sharp cone aeroshell measures structural response driven by turbulent boundary-layer pressure fluctuations. These spatiotemporal voltage measurements are segmented into fixed-duration windows and provided as input to a CNN that outputs estimates of freestream velocity (zero AoA study) or body-referenced in-plane velocity components (nonzero AoA study), from which angle of attack (AoA) is computed. The wind tunnel dataset and label preprocessing are described in Section~\ref{sec:hwt_data}, and the training/evaluation protocol is described in Section~\ref{sec:training_eval}.

\subsection{Problem formulation and target quantities}\label{sec:method_formulation}

We model the mapping from windowed piezoelectric measurements to flow state as
\begin{equation}
\bm v = \mathcal{M}_{\bm\theta}(\bm s),
\end{equation}
where \(\mathcal{M}_{\bm\theta}: \mathbb{R}^{n_s \times n_t}\rightarrow \mathbb{R}^k\), \(\bm\theta \in \mathbb{R}^m\) are trainable parameters, and \(\bm s \in \mathbb{R}^{n_s \times n_t}\) contains \(n_t\) time samples from \(n_s\) piezoelectric sensors over a fixed-duration measurement window. The velocity output \(\bm v \in \mathbb{R}^k\) is assumed approximately constant over the window.

\begin{figure}[h!]
    \centering
    \includegraphics[width=0.60\linewidth]{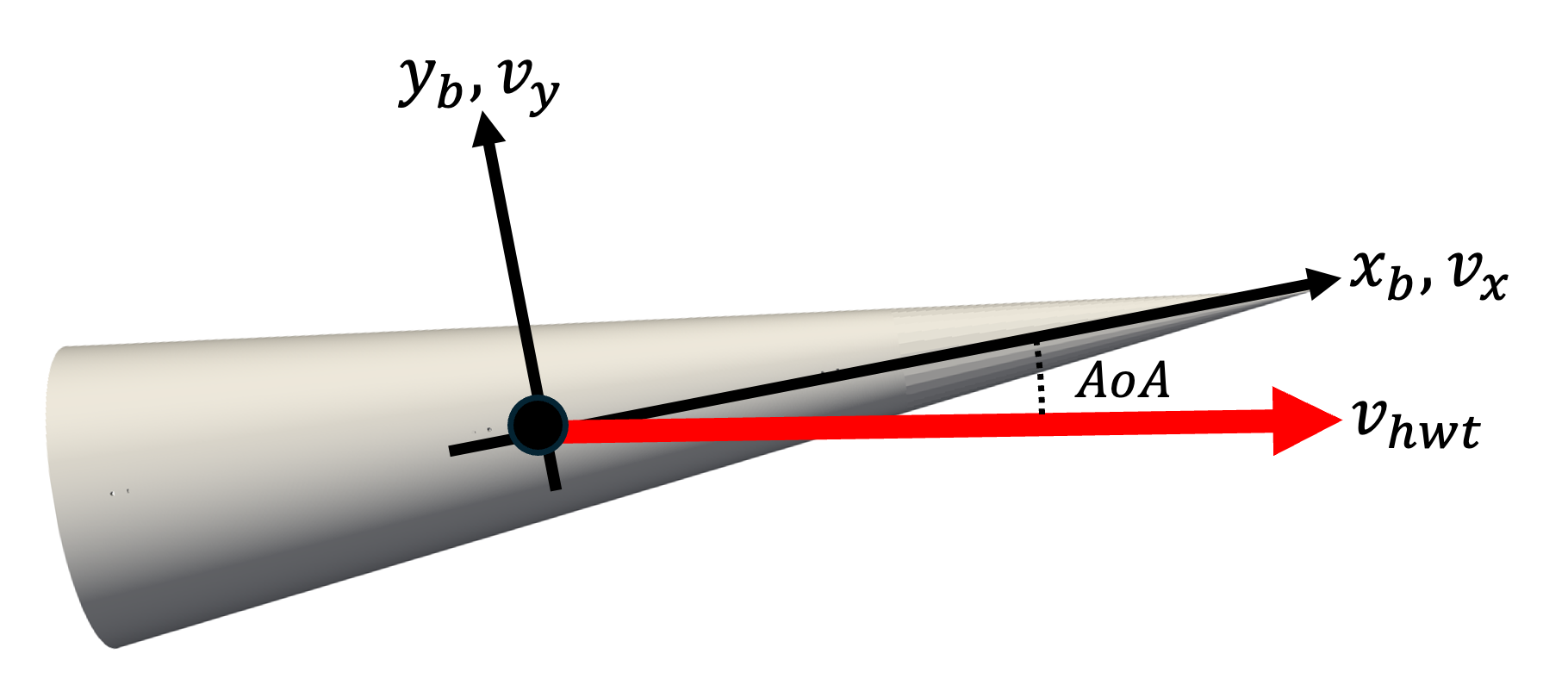}
    \caption{Body-referenced coordinate frame \((x_b,y_b)\) and estimated in-plane body-referenced velocity components \((v_x,v_y)\). The vector \(v_{hwt}\) indicates the wind tunnel flow direction and magnitude used to define the freestream reference.}
    \label{fig:bodyframe}
\end{figure}

We consider two related estimation problems. For the zero AoA study, \(k=1\) and the network estimates a reference freestream velocity label. For the nonzero AoA study, \(k=2\) and the network estimates the in-plane body-referenced velocity components \(\bm v = (v_x,v_y)\) defined in the coordinate frame shown in  Figure~\ref{fig:bodyframe}. From the estimated components, the AoA is computed as
\begin{equation}
AoA = \tan^{-1}\!\left(\frac{v_y}{v_x}\right),
\end{equation}
using the sign convention indicated in Figure~\ref{fig:bodyframe}. Although the wind tunnel configuration has stationary hardware and moving air, we use the equivalent kinematic interpretation of a body moving through air to define the body-referenced velocity components and AoA.

\subsection{CNN architecture}\label{sec:cnn_architecture}
The CNN architecture is designed to exploit the coupled spatial--temporal structure of the sensor-array measurements. Adjacent sensors experience correlated motion due to structural continuity, and the response evolves over time according to the underlying structural dynamics. Two-dimensional convolutions acting jointly over sensor index and time preserve this local structure and are well suited to learning spatiotemporal patterns (e.g., propagating wave features) that may be informative for flow-state inference.

 Similar 2D-CNN approaches have been used successfully in other wavefield inference settings. In seismology, Vantassel et al. \cite{Vantassel2022} successfully applied 2D-CNNs to high-resolution subsurface imaging using seismic wavefield surface measurements. Similarly, a comprehensive review of structural health monitoring by Jia and Li document how 2D convolutions capture damage signatures manifesting as coupled spatial-temporal features across sensor networks \cite{Jia2023}.  The 2D kernel design captures physical phenomena by learning spatiotemporal patterns that evolve coherently in time. This is consistent with wave propagation physics by preserving the notion that nearby sensors at nearby times are more correlated than distant sensors at distance times. 

The network comprises three convolutional stages with 32, 64, and 128 feature channels, respectively, yielding a model with approximately \(2\times 10^5\) trainable parameters. Each stage uses leaky-ReLU activations and group normalization to improve training stability for small batch sizes and to reduce sensitivity to batch-to-batch statistics. An adaptive pooling layer is used to accommodate different measurement window lengths without manual tuning of pooling kernel sizes.

\section{Experimental Setup}\label{sec:hwt_data}

\begin{figure}[h!]
    \centering
    \includegraphics[width=0.75\linewidth]{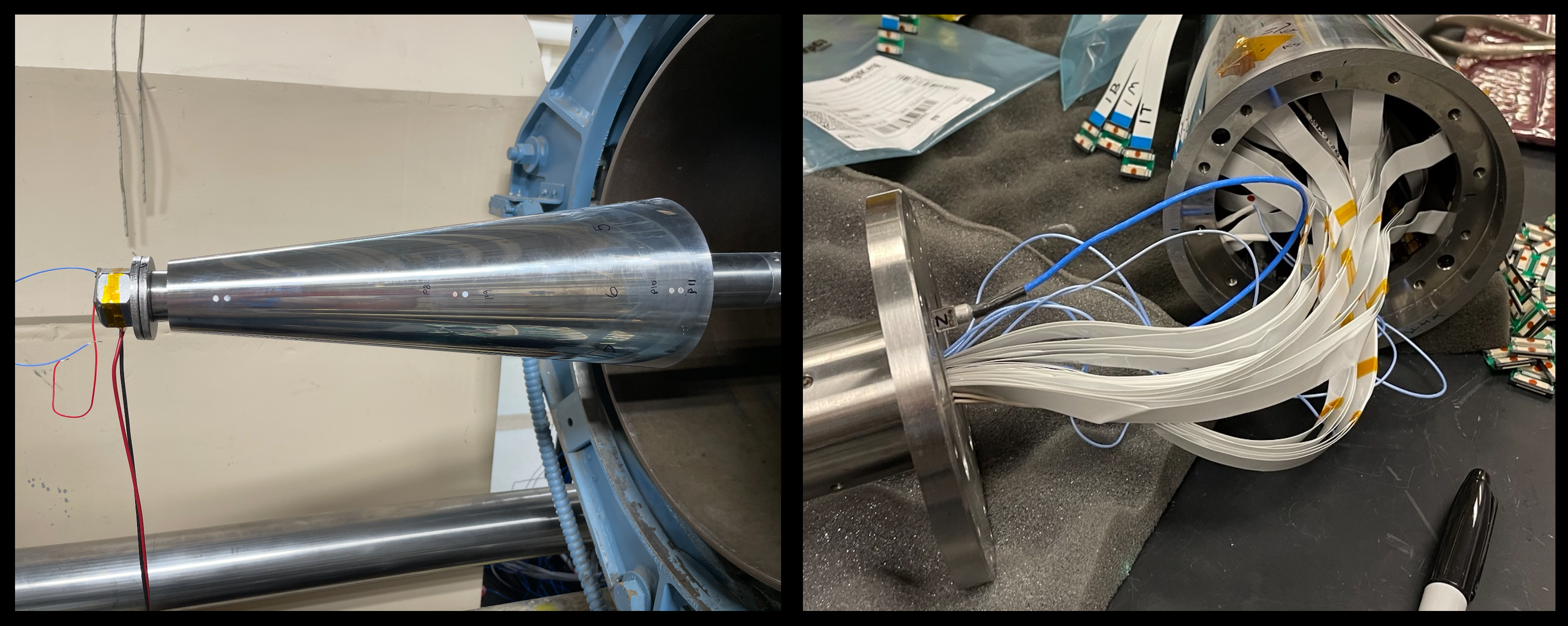}
    \caption{The sharp cone as installed in the Sandia HWT and the cabling to the piezoelectric network. The nose tip has been temporarily removed and replaced with a calibrator.}
    \label{fig:hwt-after-install}
\end{figure}

\begin{figure}[h!]
    \centering
    \includegraphics[width=0.75\linewidth]{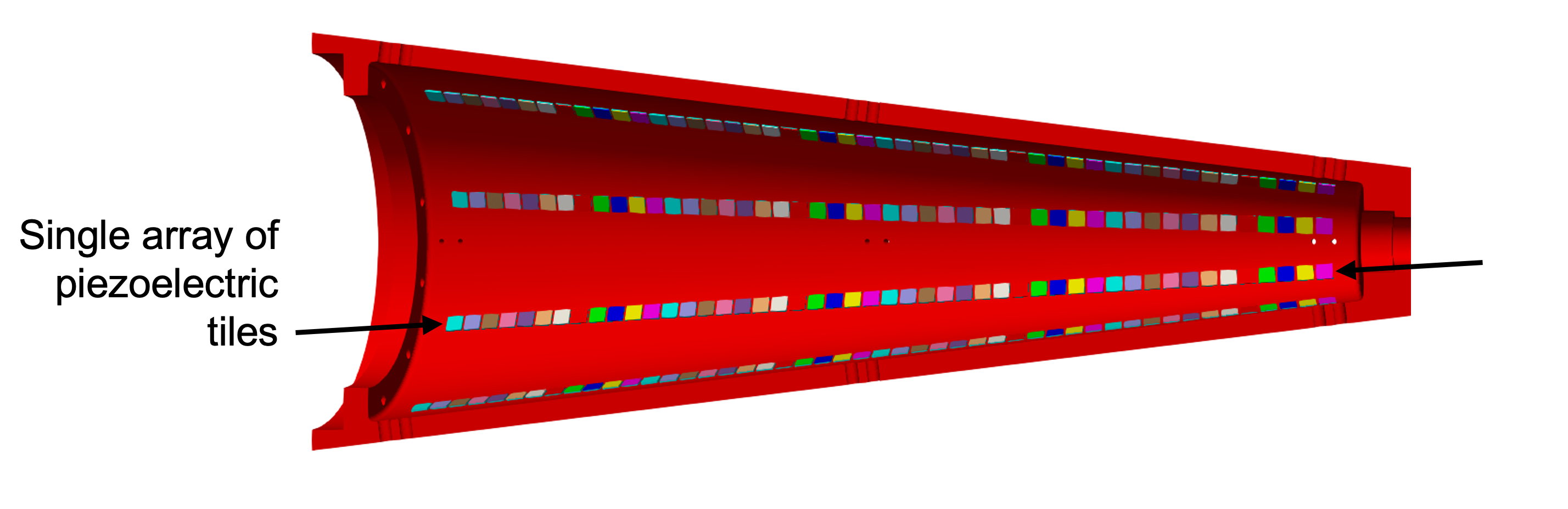}
    \caption{Diagram of the internal piezoelectric sensor layout where a single array of 48 piezoelectric sensors is highlighted. }
    \label{fig:piezo_diagram}
\end{figure}

The Sandia Hypersonic Wind Tunnel (HWT) is a conventional blowdown-to-vacuum facility capable of supporting Mach 5, 8, or 14 flows \cite{Casper2019}. The test article is a 0.517~m long, $7^{\circ}$ half-angle, sharp, stainless-steel, hollow cone with a shell thickness of 12.7~mm (Figure~\ref{fig:hwt-after-install}). This sharp cone geometry is a standard HWT test article that has been used extensively in the open literature to study hypersonic flow characteristics in the Sandia facility \cite{Casper2019}. The cone is equipped with eight arrays of piezoelectric tiles with each array comprised of 48 evenly-spaced tiles running along the length of the cone. The eight arrays are uniformly distributed around the circumference of the cone (Figure~\ref{fig:piezo_diagram}). Each piezoelectric tile measures 53~mm wide and 3~mm thick and are affixed to the cone's inner curved surface using a thin layer of epoxy. In this work, we use a single 48-sensor array with extensions to multiple arrays reserved for future work. 

During each wind tunnel run, piezoelectric voltage response and wind tunnel velocity data were recorded over a 16~s interval. Piezoelectric voltages were sampled at 250~kHz to capture the relevant structural dynamics under hypersonic flow. Tunnel velocity and AoA measurements were sampled at 100~Hz. The experiment faced challenges related to electromagnetic interference, which introduced spurious measurement noise into the measured voltage. To address this issue, advanced signal processing techniques were used to effectively remove the noise, thereby enhancing the quality and reliability of the piezoelectric data collected.  

A total of 16 runs were performed spanning Mach~5 and Mach~8 conditions, multiple Reynolds-number states (low/mid/high), and both constant and time-varying operating histories in velocity and(or) AoA. The specific HWT runs corresponding to different flow conditions, referred to simply as "runs", are detailed in Table~\ref{tbl:test_runs}. 

\begin{table}[ht]
 \centering
\caption{HWT Test Runs Description}
\begin{tabular}{|c|c|c|c|c|}
    \hline
    & \textbf{Run Label} & \textbf{Mach Number} & \textbf{Reynolds Description} & \textbf{Angle-of-Attack} \\ 
    \hline
    \centering\multirow{7}{*}{\rotatebox{90}{\textbf{Zero AoA}}} & Z-1 & 8 & Low & Hold $0^\circ$ \\
   & Z-2 & 8 & High & Hold $0^\circ$ \\
   & Z-3 & 8 & Low & Hold $0^\circ$ \\
   & Z-4 & 8 & High to Mid & Hold $0^\circ$ \\
   & Z-5 & 8 & Mid to Low & Hold $0^\circ$ \\
   & Z-6 & 5 & Low & Hold $0^\circ$ \\
   & Z-7 & 5 & Mid to Low & Hold $0^\circ$ \\
   \hline
   \centering\multirow{9}{*}{\rotatebox{90}{\textbf{Nonzero AoA}}} & NZ-1 & 8 & Mid & $0^\circ$,$\pm 8.5^\circ$,$\pm 6^\circ$,$\pm 3^\circ$\\
    & NZ-2 & 8 & Mid & $0^\circ$,$\pm 8.5^\circ$,$\pm 6^\circ$,$\pm 3^\circ$ \\
    & NZ-3 & 8 & High & $0^\circ$,$\pm 8.5^\circ$,$\pm 6^\circ$,$\pm 3^\circ$ \\
    & NZ-4 & 8 & Low & Hold $9^\circ$\\
    & NZ-5 & 8 & High & Hold $9^\circ$\\
    & NZ-6 & 5 & Low & $0^\circ$,$\pm 8.5^\circ$,$\pm 6^\circ$,$\pm 3^\circ$ \\
    & NZ-7 & 5 & Mid & $0^\circ$,$\pm 8.5^\circ$,$\pm 6^\circ$,$\pm 3^\circ$ \\
    & NZ-8 & 5 & High & $0^\circ$,$\pm 8.5^\circ$,$\pm 6^\circ$,$\pm 3^\circ$ \\
    & NZ-9 & 5 & High & Hold $9^\circ$\\
    \hline
\end{tabular}
\label{tbl:test_runs}
\end{table}

The reference freestream velocity in the HWT is not measured with a dedicated velocity sensor; instead it is inferred from a calibrated nozzle Mach number and tunnel thermodynamic conditions. The freestream Mach number is obtained through nozzle calibration, the freestream static temperature is computed from the measured or facility-calculated stagnation temperature, and standard gas properties are then used to convert these quantities into a freestream velocity estimate. Based on facility documentation, this inferred velocity carries an approximately 7\% systematic (bias/epistemic) uncertainty. This value should be interpreted as a bound on absolute velocity accuracy, not as run-to-run random scatter or a stated percentile/standard deviation of the time-resolved signal.

The time-resolved reference velocity also exhibits high-frequency fluctuations. For use as a supervised learning label, we apply a low-pass filter to the facility-provided velocity time history to obtain a smoothed reference trajectory. Throughout this paper, CNN performance is quantified relative to this low-pass filtered reference velocity. Accordingly, the reported CNN-estimated errors represent agreement with the facility's best available time-resolved velocity estimate within this experimental campaign; they do not imply absolute freestream velocity accuracy below the facility's stated systematic uncertainty. Unlike velocity, the measured AoA is treated as known and did not require filtering.

\begin{figure}[h!]
    \centering
    \includegraphics[width=0.7\linewidth]{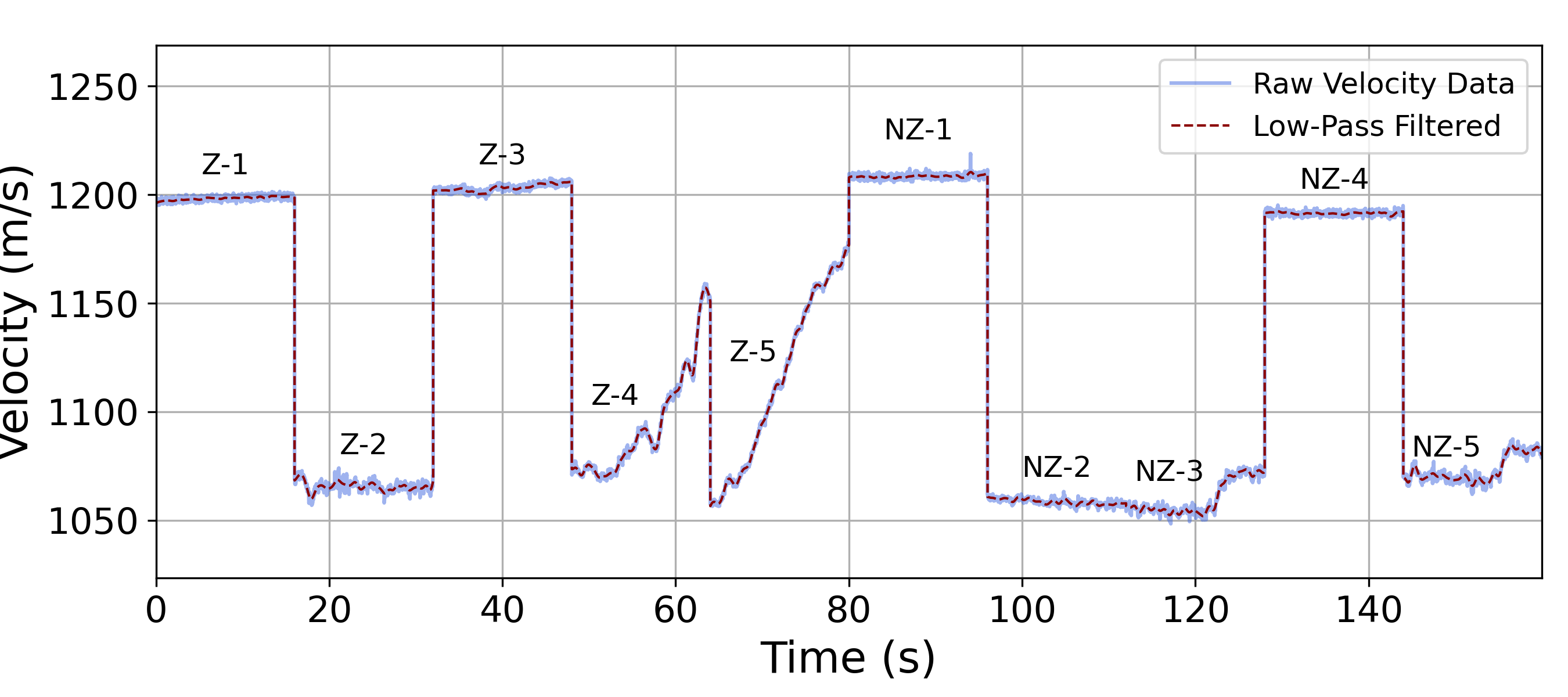}
    \caption{The raw reference wind tunnel velocity estimate (blue) and the low-pass filtered reference velocity (dashed-red) for Mach~8 runs.}
    \label{fig:mach_8_data}
    
    \vspace{0.5cm} 
    
    \includegraphics[width=0.7\linewidth]{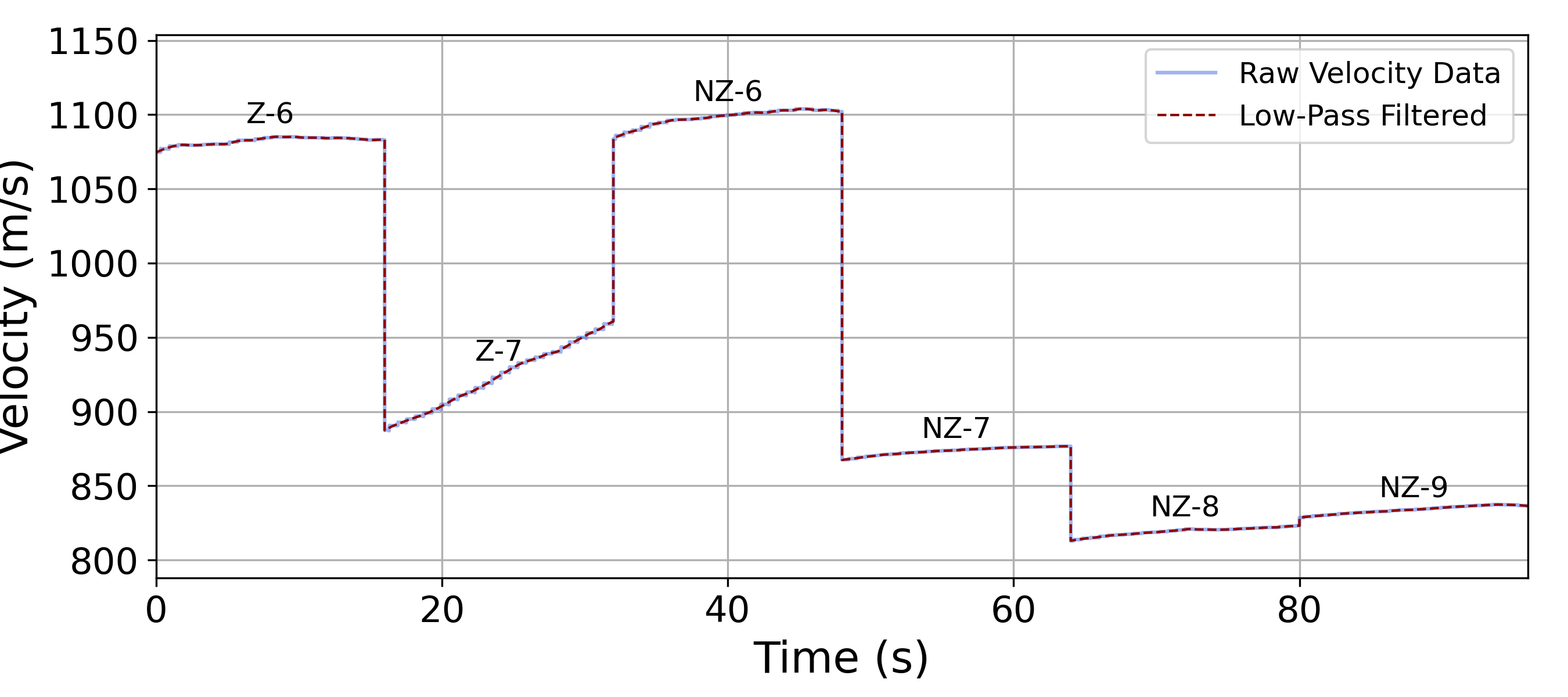}
    \caption{The raw reference wind tunnel velocity estimate (blue) and the low-pass filtered reference velocity (dashed-red) for Mach~5 runs.}
    \label{fig:mach_5_data}
\end{figure}

Figures~\ref{fig:mach_8_data} and~\ref{fig:mach_5_data} show the raw and filtered velocities for Mach~8 and Mach~5 runs, respectively. Note that each run was not necessarily tested sequentially as suggested by the time stamps in these figures. The present discontinuities are an artifact of plotting and represent a delineation between different HWT runs. The use of a continuous range of time serves as a convenient way to visualize the different run velocities (or AoA), the absolute and relative length of each run, and the total amount of available training and test data. This convention is used for the remainder of the paper. 

\begin{figure}[h!]
    \centering
    \includegraphics[width=0.70\linewidth]{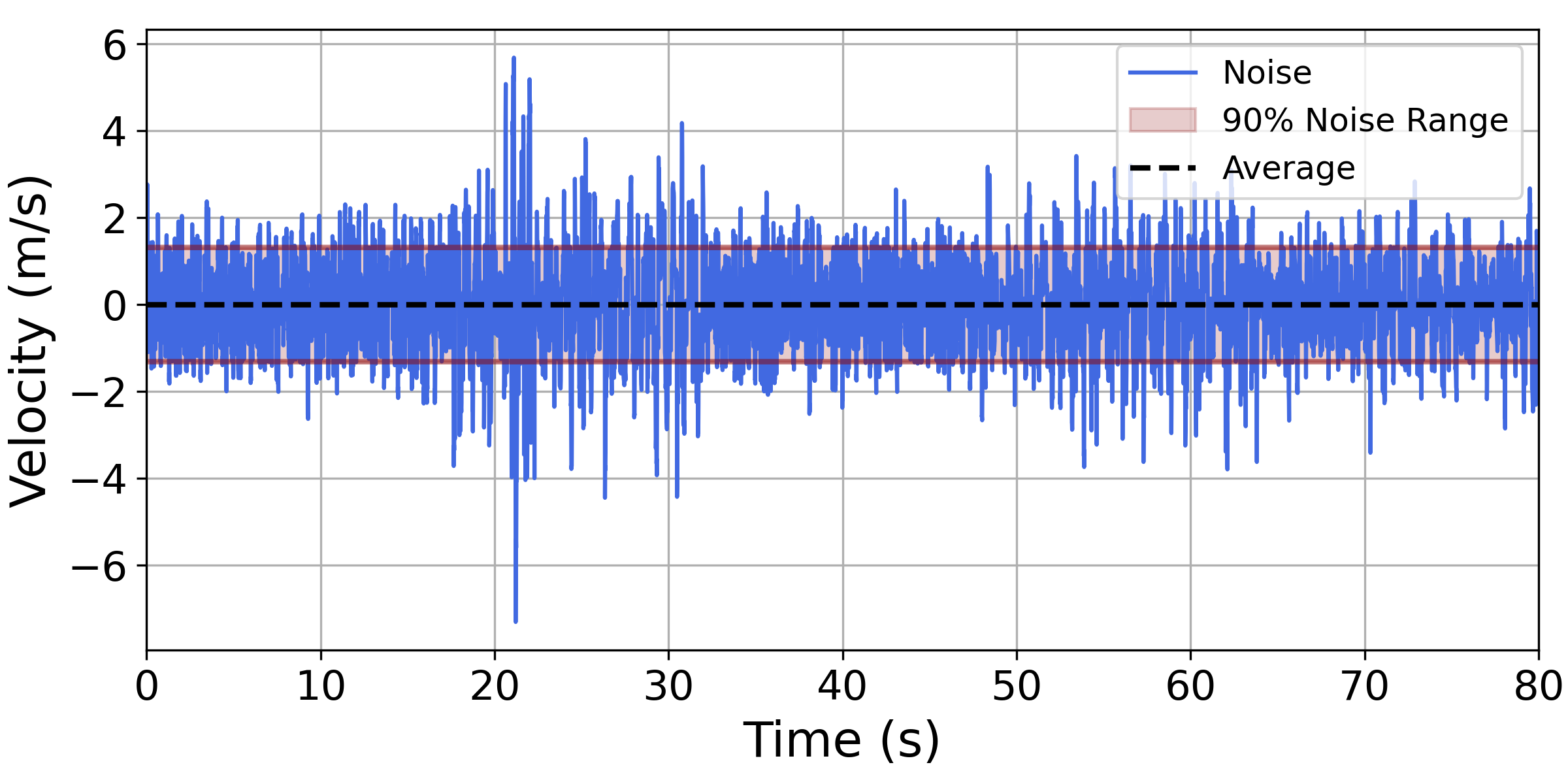}
    \caption{Mach~8 velocity high-frequency fluctuations (blue), computed as the facility velocity estimate minus the low-pass filtered reference. The mean (black dashed) and 90th percentile bounds (red) are indicated.}
    \label{fig:noise}
    \vspace{0.5cm} 

    \includegraphics[width=0.70\linewidth]{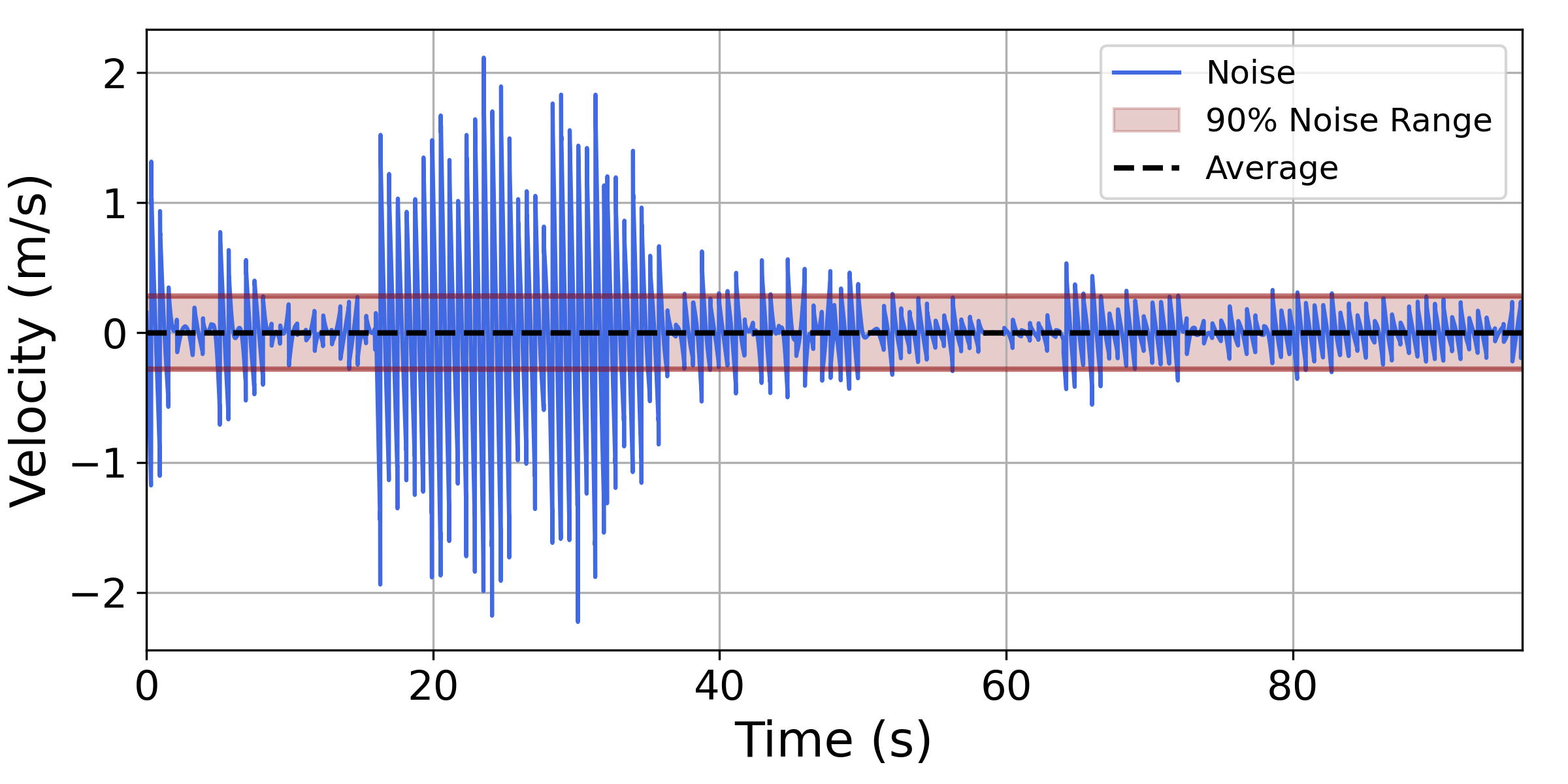}
    \caption{Mach~5 velocity high-frequency fluctuations (blue), computed as the facility velocity estimate minus the low-pass filtered reference. The mean (black dashed) and 90th percentile bounds (red) are indicated.}
    \label{fig:noise_m5}
\end{figure}

Subtracting the filtered signal from the raw reference velocity estimate yields an empirical characterization of the high-frequency fluctuation level. For Mach~8 runs, these fluctuations are approximately zero mean with 90\% of samples within \(\pm~1.3~\textrm{m/s}\) of the filtered velocity (Figure~\ref{fig:noise}). For Mach~5 runs, the velocity estimate exhibits smaller fluctuations but appears piecewise due to the measurement technique employed in the facility (Figure~\ref{fig:noise_m5}). 

In what follows, ``reference velocity'' denotes the facility-provided freestream velocity estimate after low-pass filtering; all reported velocity errors are computed relative to this reference.

\subsection{Test-training data partitioning methodology}\label{sec:training_eval}

To ensure the integrity of network training and performance evaluation, we implemented a \emph{center-block holdout} strategy for separating the time-series measurements (piezoelectric signals and corresponding velocity/AoA labels) into training, validation, and test datasets. This approach prevents leakage from training/validation data into the temporally withheld ``unseen'' test interval used to evaluate performance, while still allowing overlapping windows during training to enhance the temporal representation. In this framework, we used the training dataset to optimize the trainable parameters of the CNN, the validation dataset to check model generalization and assess convergence, and finally the test dataset to evaluate the performance of the CNN on unseen data. 

The data separation process begins by dividing the dataset by wind tunnel run, where each run is further split into training, validation, and test subsets. The test set is a contiguous block selected from the temporal center of each run, while validation data are selected near the beginning and end of each run. This approach results in temporal separation between the test and training-validation data and ensures no training-validation data leaks into the test data. The proportions of validation and test data within each run are both set to 10\% of the run duration (1.6~s per run), while the remaining 80\% is allocated to training.

\begin{figure}[h!]
    \centering
    \includegraphics[width=1\linewidth]{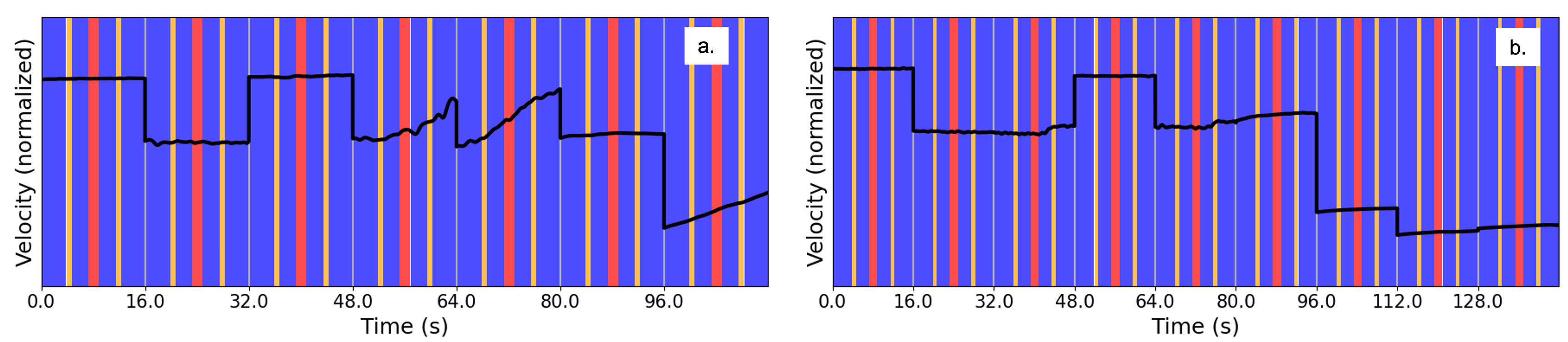}
    \caption{Center-block holdout data partitioning strategy for all zero AoA runs (a) and nonzero AoA runs (b). Red and orange regions indicate test and validation data, respectively, withheld before applying overlapping windows to training data (blue). Each HWT run is delineated by the white line. By performing partitioning prior to applying overlapping windows, we ensure complete independence between the training/validation and test sets. }
    \label{fig:test_train_partition}
\end{figure}

The zero AoA and nonzero AoA training-test partitions are shown in Figure~\ref{fig:test_train_partition}, where blue represents the training data, red indicates the test data, orange indicates the validation data, white vertical lines delineate the HWT runs, and the black curve is the normalized reference freestream velocity of the wind tunnel. 

Once the data is separated into training, validation and test subsets, we divide the time series data into overlapping windows. Each window spans 16~ms, corresponding to 4000 voltage samples, and is assigned a label using the mean of the available reference velocity/AoA measurements over that interval (scalar velocity or a 2D velocity vector for nonzero AoA). Since the 16~ms window is slightly longer than the 10~ms sampling period of the tunnel velocity measurements, this averaging can introduce additional error when conditions vary rapidly. 
This window length represents a hyperparameter that could be optimized in future work; here, we hold it constant for all studies. 

When training the network, consecutive windows overlap by 80\%, advancing 3.2~ms between successive samples. For reporting performance metrics, however, we evaluate using non-overlapping windows for both training and test sets. During testing, windows advance sequentially without overlap, providing an assessment that avoids inflated sample counts due to overlap and yields one prediction per independent time interval.

For nonzero AoA, the measured AoA and freestream velocity are converted into an equivalent body-referenced velocity vector \((v_x,v_y)\) using the convention of Figure~\ref{fig:bodyframe}, and these components are used as the CNN training labels.

\section{Velocity and AoA Estimation Experimental Validation}\label{sec:validation}

\subsection{Reference freestream velocity estimation under zero AoA conditions}\label{sec:zeroaoa_free}
The zero AoA velocity estimation problem serves as our baseline study, isolating reference freestream velocity inference from directional velocity sensing to assess the CNN's ability to invert piezoelectric-based structural response measurements into flow conditions. The seven zero AoA wind tunnel tests (denoted Z-1 through Z-7 in Table~\ref{tbl:test_runs}) span both steady-state and continuously varying flow regimes at Mach~5 and Mach~8. All results in this section use a \emph{center-block holdout} split: for each run, a contiguous block of samples at the temporal center is withheld for testing, and the remaining (disjoint) early and late portions of the run are used for training and validation.

\begin{figure}[h!]
    \centering
    \includegraphics[width=0.70\linewidth]{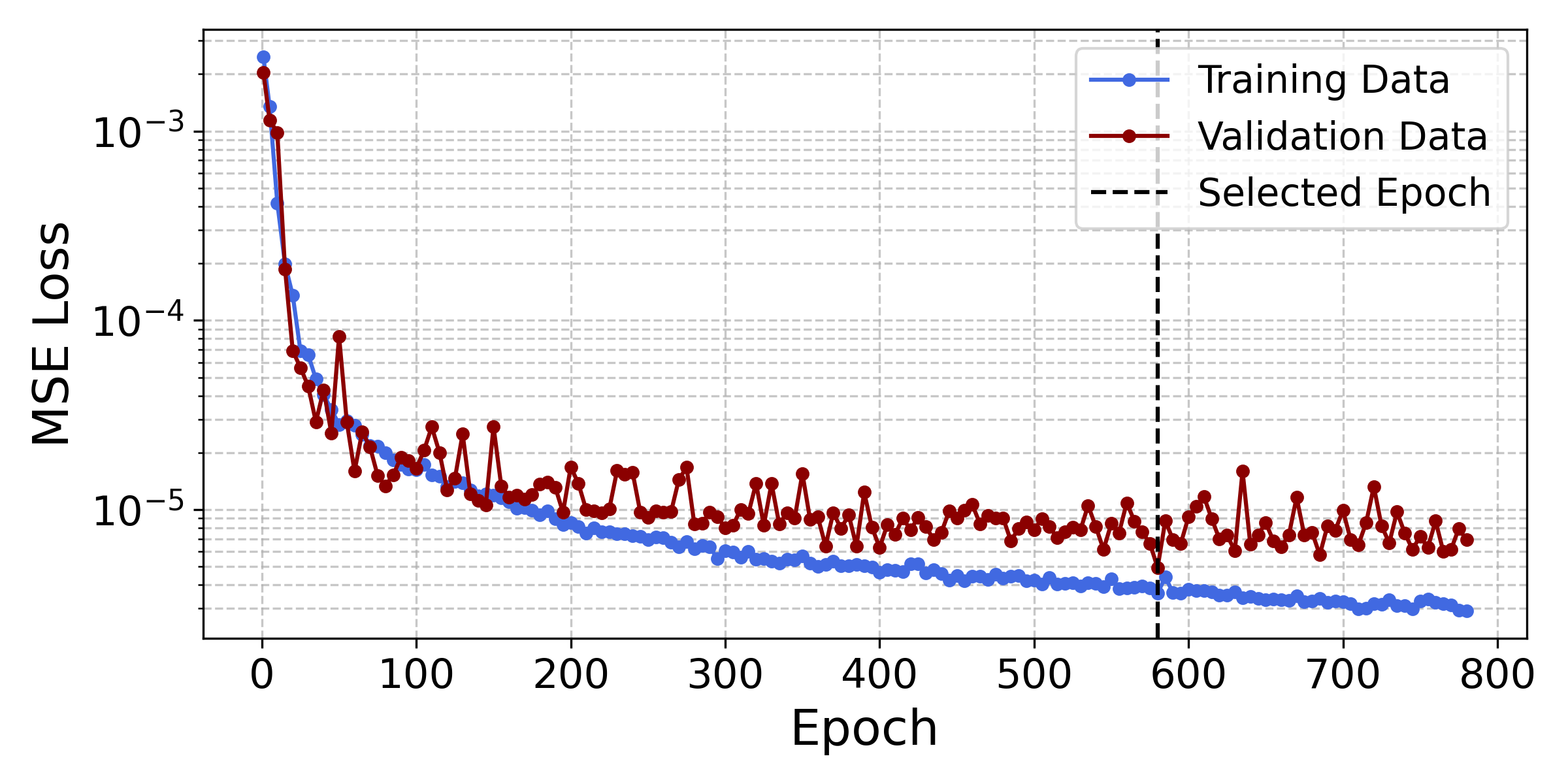}
    \caption{The MSE loss function using the training (blue) and validation (red) datasets as a function of the CNN training epoch. The black dashed line represents the selected stopping point corresponding to minimum validation loss. }
    \label{fig:training_loss}
\end{figure}

Network training convergence characteristics appear in Figure~\ref{fig:training_loss}, which plots the MSE loss versus epoch for both the training and validation datasets. The training loss exhibits the expected monotonic decrease, while the validation loss follows a similar trajectory with increased variance. The observed plateau indicates that the model has extracted the dominant learnable patterns from the available data without continued improvements in generalization. We adopt the model checkpoint at minimum validation loss (indicated by the dashed line) with a patience value of 200 epochs.

Performance is quantified using the mean and 90th-percentile absolute velocity error, \(e = |v_{ref} - v_{est}|\), where \(v_{ref}\) is the low-pass filtered facility referenced velocity measurement and \(v_{est}\) is the CNN estimate. The mean reflects typical performance, while the 90th percentile captures tail behavior that is not dominated by rare outliers.

\begin{figure}[ht]
    \centering
    \includegraphics[width=0.70\linewidth]{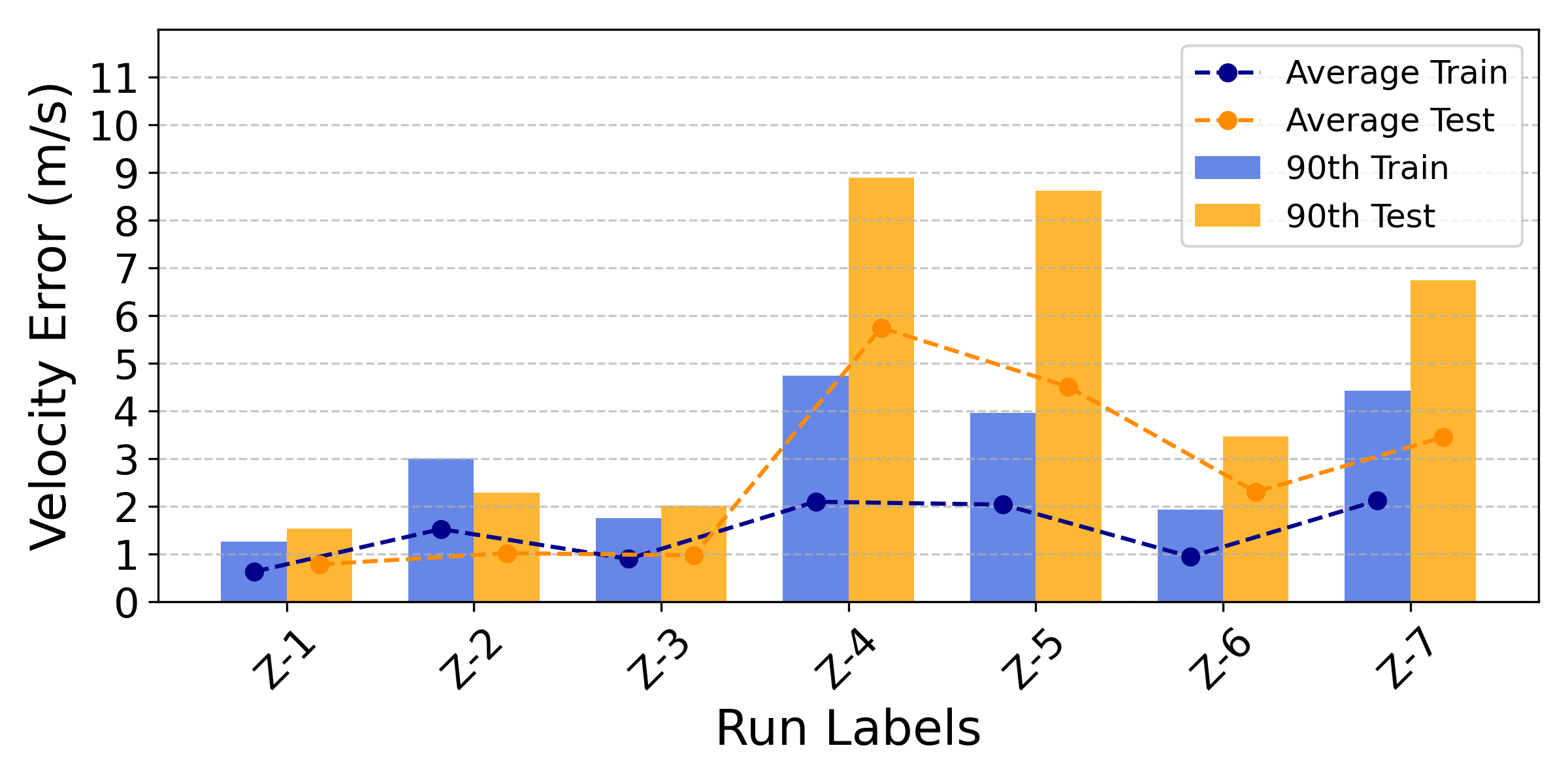}
        \caption{Raw CNN velocity estimation errors for zero AoA cases by run showing mean (dashed-lines) and 90th-percentile (bars) for training (blue) and test (orange) data. Comparable test and training performances confirm generalization. Errors were substantially smaller during constant tunnel conditions experiments (Z-1, Z-2, Z-3, Z-6).}
        \label{fig:zeroAoAExpResultsArr1}
\end{figure}

Figure~\ref{fig:zeroAoAExpResultsArr1} reports the mean and 90th-percentile errors for each run, decomposed by dataset (training vs.\ test). Training and test performance are comparable, indicating that the model generalizes well to the withheld test interval within each run rather than simply memorizing the training windows. Run-to-run differences are more pronounced. Runs Z-1, Z-2, Z-3, and Z-6, all conducted at constant tunnel conditions, achieve excellent performance (mean error <2.5~m/s; 90th-percentile error <3.5~m/s). In contrast, runs Z-4, Z-5, and Z-7, the continuously varying velocity cases, exhibit substantially higher errors (mean: 2--6~m/s; 90th: 4--9~m/s). Because the split is performed within each run (i.e., training and testing occur under the same  tunnel condition for a given run), this pattern indicates that the dominant challenge is not estimating velocity under steady, time-invariant conditions, but maintaining accurate predictions when the tunnel velocity varies continuously in time. Under constant conditions, the withheld center block is statistically similar to the training blocks, so the model is trained on highly representative examples. Under continuously varying conditions, the withheld center block contains intermediate velocities that are less well represented in the remaining  training segments, leading to increased variance and larger tail errors (consistent with fewer samples per velocity level when acceleration is higher).

\begin{figure}[ht]
    \centering
    \includegraphics[width=0.70\linewidth]{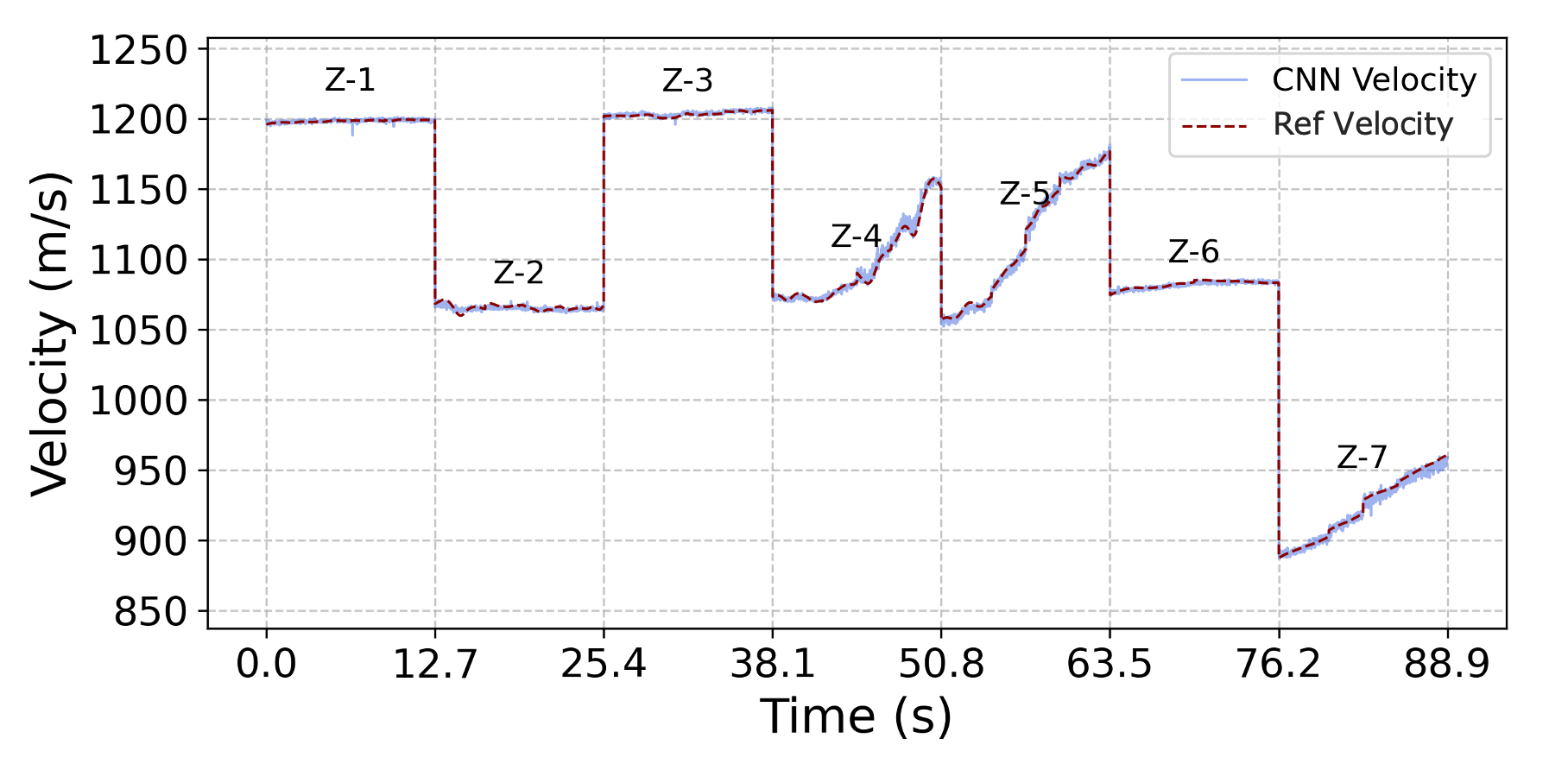}
        \caption{Time history of reference velocity (red-dashed) and raw CNN estimates (blue) for training data across seven runs. Horizontal segments indicate constant tunnel conditions; sloped segments indicate varying tunnel conditions. }
        \label{fig:raw_error_training_2_zeroaoa}

     \includegraphics[width=0.70\linewidth]{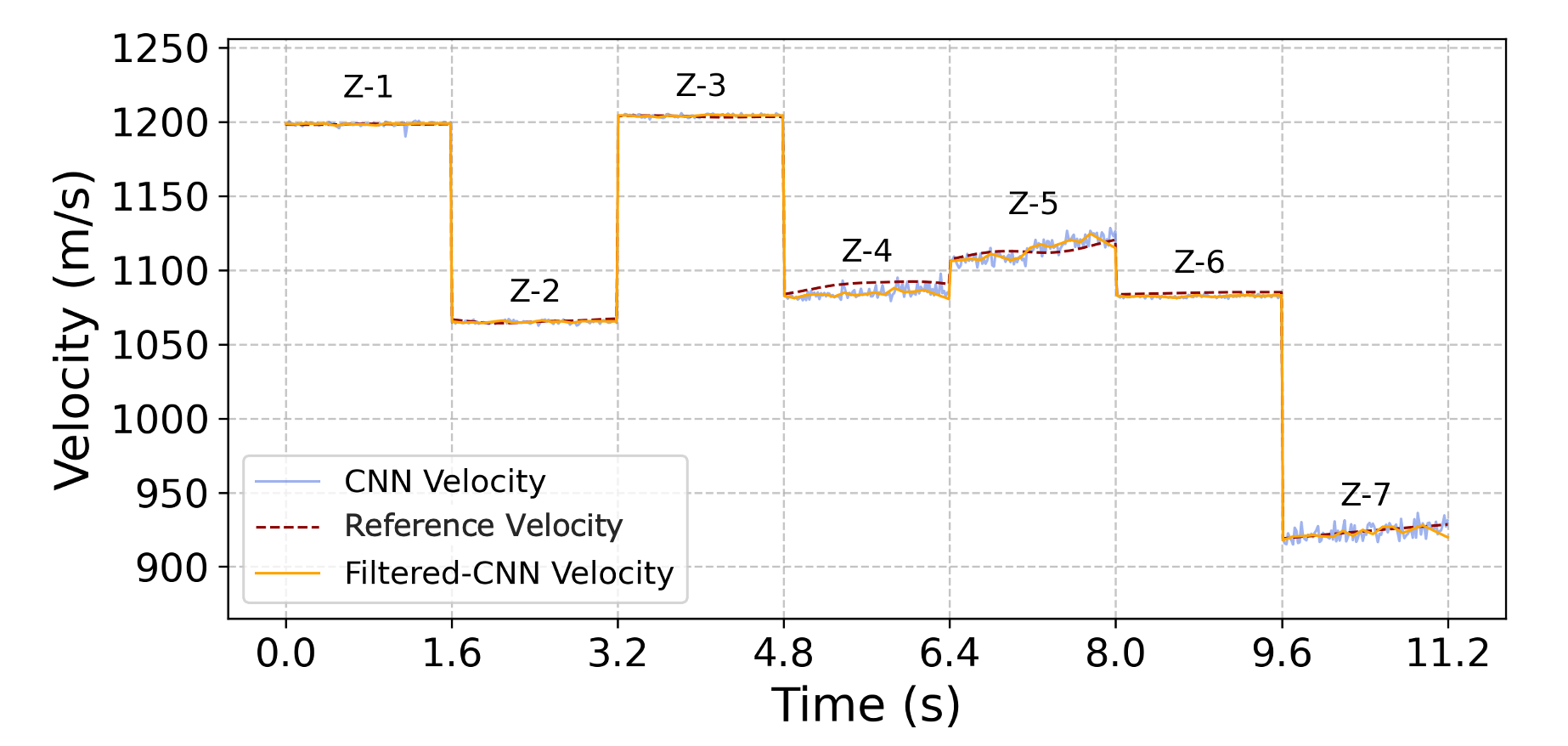}
        \caption{Time history of reference velocity (red-dashed), raw CNN estimates (blue), and median-filtered estimates (orange) for test dataset. Median filtering reduces variance, while preserving tracking accuracy across all seven zero AoA test runs.}
        \label{fig:raw_error_test_2_zeroaoa}
\end{figure}

The time-series plots in Figures~\ref{fig:raw_error_training_2_zeroaoa} and \ref{fig:raw_error_test_2_zeroaoa} make this steady/transient distinction visually apparent. These figures show the reference velocity (red) alongside CNN estimates (blue) for the training and test datasets, respectively. During constant flow conditions (Z-1, Z-2, Z-3, Z-6), the estimates remain stable and track the ground truth closely with minimal high-frequency oscillations. During continuously varying conditions (Z-4, Z-5, Z-7), the estimates become noticeably noisier, fluctuating about the reference trajectory. Importantly, these fluctuations remain approximately centered on the ground truth, indicating that the dominant transient error mode is increased variance rather than a large systematic bias.

This observation motivates the introduction of a sliding median filter as a post-processing step. By computing the median over a window of 6 consecutive CNN estimates, we can suppress high variance while preserving the accurate mean trajectory, thus transforming noisy predictions into smooth, directly usable velocity estimates. When implemented using non-overlapping windows to avoid correlated estimates, this reduces the effective refresh rate from 62.5~Hz (16~ms sampling) to 10.4~Hz.   

\begin{figure}[h!]
    \centering
    \includegraphics[width=0.70\linewidth]{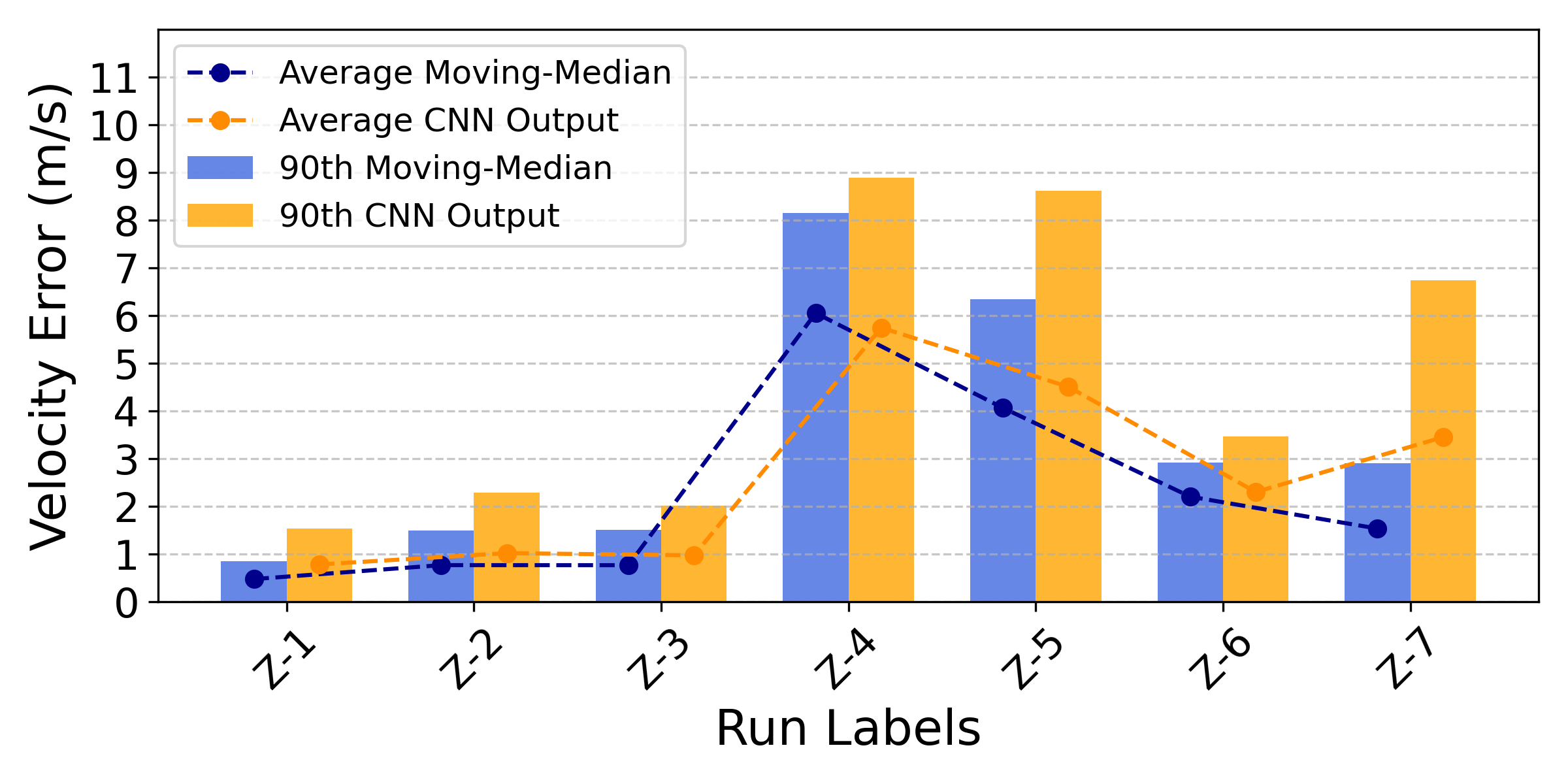}
    \caption{Raw CNN output (orange) and moving-median filtered (blue) velocity estimation errors for zero AoA cases by run, showing mean (dashed-lines) and 90th-percentile (bars). Moving-median results in  decreases in 90th-percentile errors across all runs, but most significantly for continuously-varying tunnel conditions (Z-4, Z-5, Z-7).}
    \label{fig:compare_filter_zeroAoA}
\end{figure}

Applied to the test data with a 6-sample window, the median-filtered estimates (shown in orange in Figure~\ref{fig:raw_error_test_2_zeroaoa}) reduce the variance while preserving accuracy. The smoothed trace follows the reference velocity (i.e. low-pass filtered reference freestream velocity) more closely, resulting in estimation improvements across all runs. 
Figure~\ref{fig:compare_filter_zeroAoA} directly compares the raw CNN output against median-filtered estimates on the test dataset, quantifying the improvement achieved through post-processing. The enhancement is evident across all seven runs. On average, 90th percentile velocity errors decrease approximately 1.3~m/s, with the most substantial improvements occurring in runs Z-5 and Z-7, where transient flow conditions had produced the highest raw variance. This systematic improvement confirms that temporal filtering successfully suppresses unbiased noise without degrading the underlying estimation accuracy. However, the moving-median does little to improve remaining bias in the estimates.  

\subsubsection{Differential data requirements for steady vs transient conditions for zero AoA flow}\label{sec:zeroaoa_free_kfold}

\begin{figure}[h!]
    \centering
    \includegraphics[width=1.0\linewidth]{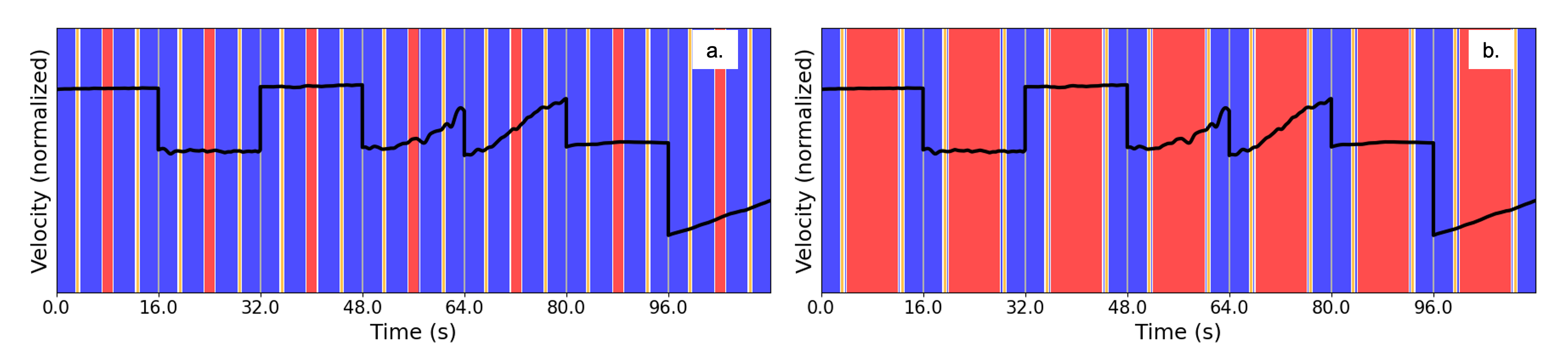}
    \caption{Train-test-validation data partitioning strategy for all zero AoA runs for center-block-holdout study where a) 10\% and b) 50\% test data is withheld from training.  Red and orange regions indicate test and validation data, respectively }
    \label{fig:kfold_testtrain}
\end{figure}

To assess training-data requirements, we performed a \emph{center-block holdout} study. For each wind tunnel run Z-1 through Z-7, a contiguous block of samples located at the temporal center of the run was withheld for testing, while the two remaining (disjoint) blocks at the beginning and end of the run were used for training and validation. The held-out block size was increased from 10\% to 50\% of each run in 10\% increments (equivalently, the available training plus validation fraction decreased from 90\% to 50\%). This blocked split reduces leakage from temporally adjacent samples and probes the model’s ability to estimate flow conditions over an unseen time interval within the same run. Figure~\ref{fig:kfold_testtrain} presents the 10\% and 50\% withheld test data cases for clarity.

\begin{figure}[h!]
    \centering
    \includegraphics[width=1.0\linewidth]{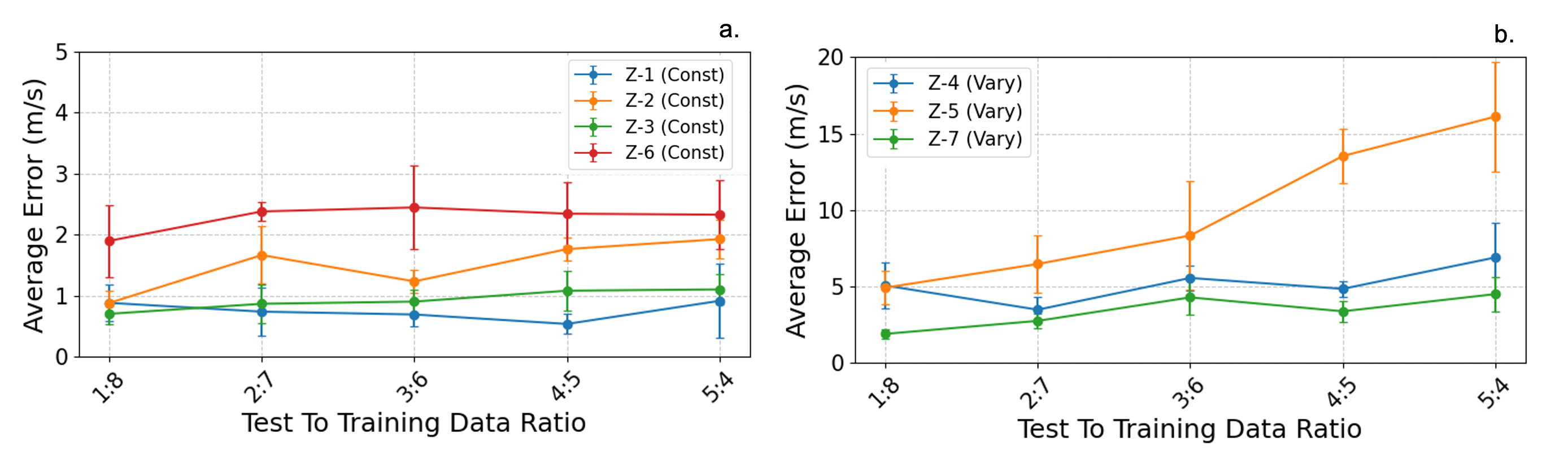}
 \caption{Center-block holdout study for zero AoA velocity estimation: a temporally centered contiguous block (10--50\%) is withheld from each run and used for testing. Points show mean absolute error over the withheld block; error bars denote $\pm 1\sigma$ across five retrainings.}
    \label{fig:kfold_zeroaoa}
\end{figure}

For each holdout fraction, we retrained the CNN five times using different random initializations and data shuffles to quantify optimization stochasticity. Figure~\ref{fig:kfold_zeroaoa} reports the mean absolute velocity estimation error for each run (computed over the withheld center block), with error bars indicating $\pm 1$ standard deviation across the five retrains. Results are grouped into (a) constant-condition runs (Z-1, Z-2, Z-3, Z-6) and (b) continuously varying-condition runs (Z-4, Z-5, Z-7).

Across runs, it appears the dominant trend is that performance is governed by how similar (i.e., correlated) the withheld center interval is to the remaining training data from the same run. For the constant-condition cases (Fig.~\ref{fig:kfold_zeroaoa}a), tunnel conditions are stationary, so the response statistics in the withheld block closely match those in the two outer training blocks. Consequently, increasing the withheld fraction from 10\% to 50\% produces no systematic rise in error and only modest retrain-to-retrain variability, indicating that accurate velocity estimation can be learned from substantially reduced data under steady conditions.

In contrast, the transient runs (Fig.~\ref{fig:kfold_zeroaoa}b) generally show increasing error as the withheld block grows. Here, enlarging the center holdout removes a larger portion of the run corresponding to intermediate velocities, decreasing the overlap between velocities represented in the remaining training blocks (early/late portions) and those present in the withheld test block. Runs Z-4 and Z-7 degrade only mildly, whereas Z-5 degrades substantially, consistent with its more rapid acceleration (fewer samples per instantaneous velocity), which further reduces the amount of training data that is representative of the withheld interval. Overall, these results suggest that the CNN can interpolate to unseen time segments, but accuracy deteriorates when the withheld interval becomes less supported by the remaining training data.

\subsection{Velocity component and AoA estimation under nonzero AoA conditions}\label{sec:aoa}

Having established the CNN's capability for freestream velocity estimation at zero AoA, we now consider the more challenging case of nonzero AoA conditions. Here, the CNN is trained to estimate both components of the 2D body-referenced velocity vector (Figure~\ref{fig:bodyframe}) and, from these components, the AoA and velocity magnitude, using the same array of 48 piezoelectric tiles and the same input window size as in the zero-AoA study.

For this investigation, we use data from nine nonzero AoA wind tunnel runs (denoted NZ-1 through NZ-9 in Table~\ref{tbl:test_runs}). As in Section~\ref{sec:zeroaoa_free}, all results use a \emph{center-block holdout} split within each run: a contiguous block at the temporal center is withheld for testing, and the remaining (disjoint) early and late portions of the run are used for training and validation. Performance is quantified using the freestream velocity magnitude (the low-pass filtered reference freestream velocity defined in Section~\ref{sec:hwt_data}) and AoA.

\begin{figure}[h!]
    \centering
    \includegraphics[width=0.70\linewidth]{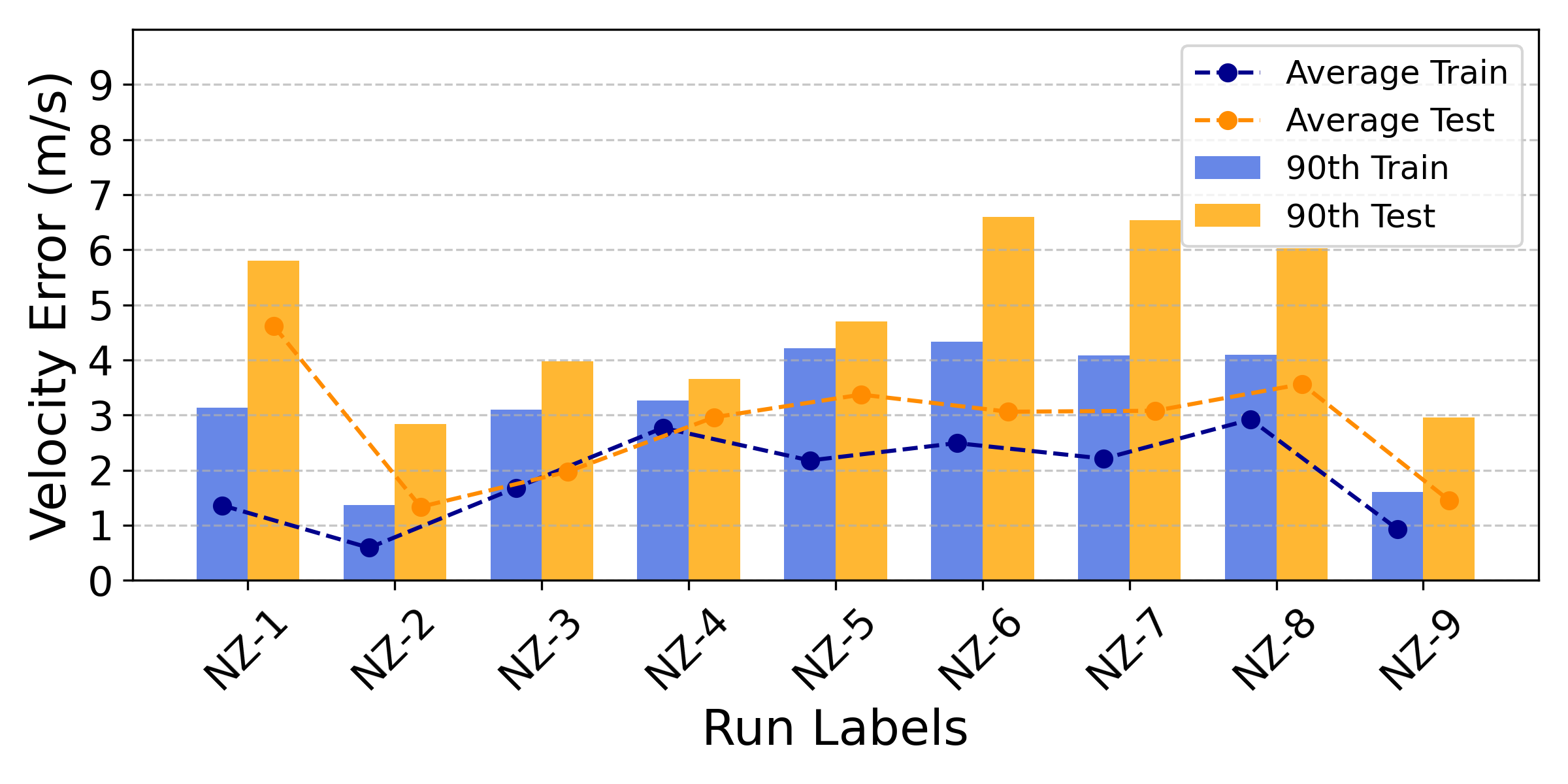}
        \caption{Raw CNN velocity magnitude estimation errors for nonzero AoA cases by run showing mean (dashed-lines) and 90th percentile (bars) for training (blue) and test (orange) data. Comparable test and training performances confirm generalization.}
        \label{fig:nonzero_compare_mag_testtrain}
    
    \vspace{0.5cm} 
    \includegraphics[width=0.70\linewidth]{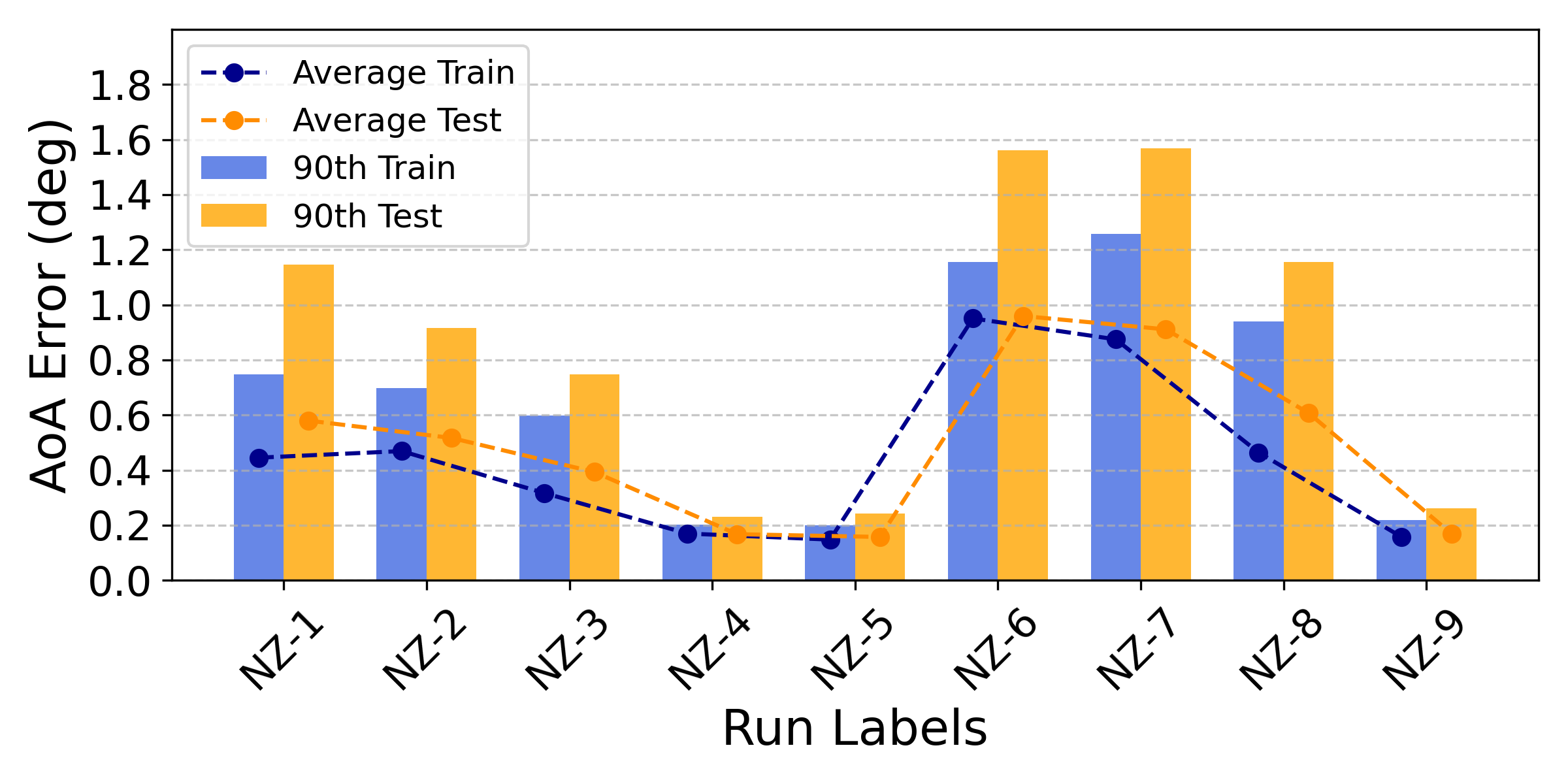}
        \caption{Raw CNN AoA estimation errors by run showing mean (dashed-lines) and 90th percentile (bars) for training (blue) and test (orange) data. Comparable test and training performances confirm generalization. Constant AoA runs (NZ-4,5,9) achieve lower errors than varying-AoA runs. }
        \label{fig:nonzero_compare_aoa_testtrain}
\end{figure}

Following the methodology established in the zero AoA study, we report mean and 90th-percentile absolute errors, decomposed by run and by dataset (training vs.\ test). Velocity magnitude errors are shown in Figure~\ref{fig:nonzero_compare_mag_testtrain}, while AoA errors are shown in Figure~\ref{fig:nonzero_compare_aoa_testtrain}. Training and test performance are comparable, indicating that the model generalizes well to the withheld (temporally centered) test interval within each run rather than simply memorizing the training windows.

Decomposing performance by run reveals systematic dependence on specific wind tunnel conditions. Figures~\ref{fig:nonzero_compare_mag_testtrain} and~\ref{fig:nonzero_compare_aoa_testtrain} show that constant AoA runs (NZ-4, NZ-5, NZ-9) achieve substantially lower errors than runs in which AoA varies continuously (NZ-1, NZ-2, NZ-3, NZ-6, NZ-7, NZ-8). Because the split is performed within each run, this steady/transient separation suggests that estimation is most reliable when the withheld center interval is statistically similar to the remaining training data. This pattern replicates our zero AoA observations, suggesting that limited training data at specific AoA-velocity instances penalizes performance and(or) AoA-velocity is inherently more challenging to estimate during transient changes in flow. We note that the nonzero-AoA dataset does not include time-varying velocity profiles tested in the zero AoA study, since such combinations of AoA and rapidly changing velocity could generate loads outside the intended operating envelope of the wind tunnel.

\begin{figure}[h!]
    \centering
    \includegraphics[width=0.70\linewidth]{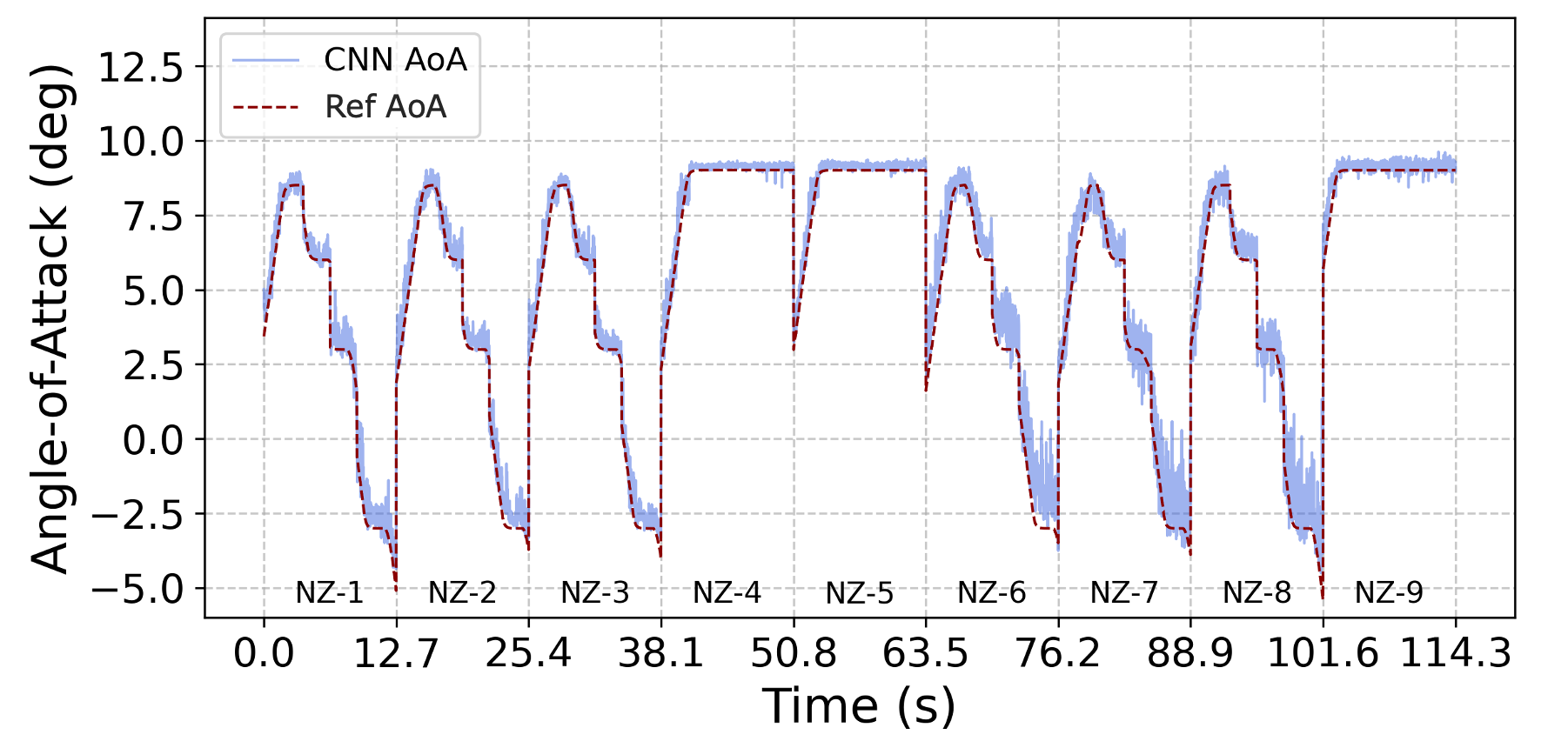}
        \caption{Time history of reference AoA (red-dashed) and raw CNN estimates (blue) for training data across nine runs. Horizontal segments indicate constant AoA conditons; sloped segments indicate varying AoA conditions. }
        \label{fig:raw_error_train_2_aoa}

\end{figure}

\begin{figure}[h!]
    \centering
    \includegraphics[width=0.70\linewidth]{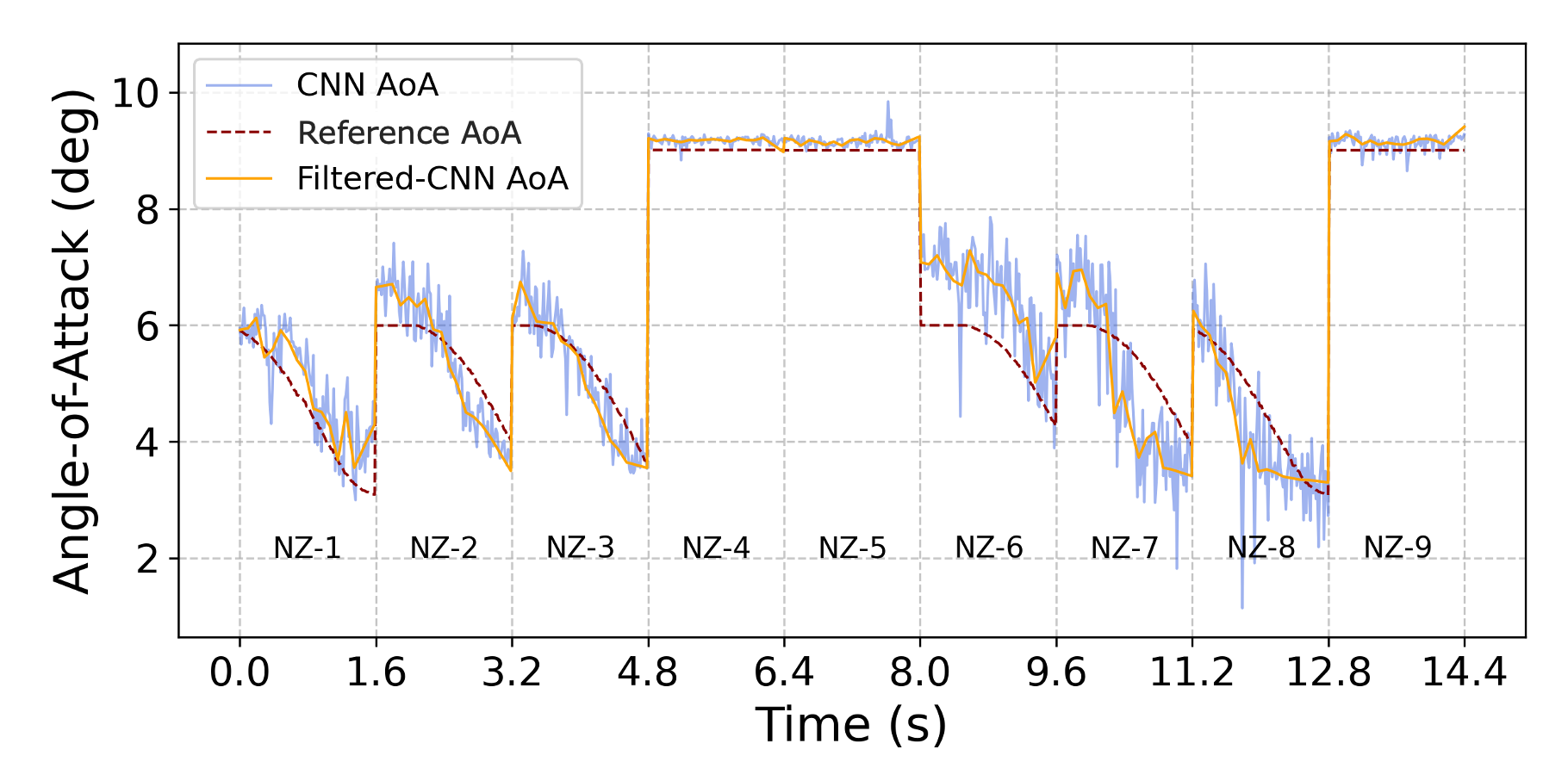}
        \caption{Time history of reference AoA (red-dashed), raw CNN estimates (blue), and median-filtered estimates (orange) for test dataset. Median filtering reduces transient-induced variance, while preserving tracking accuracy across all nine nonzero AoA test runs. }
        \label{fig:raw_error_test_2_aoa}

\end{figure}

To visualize how the CNN performs moment-by-moment rather then in aggregate, Figures~\ref{fig:raw_error_train_2_aoa} and  \ref{fig:raw_error_test_2_aoa} present the complete AoA estimation time-series. In the training data (Figure~\ref{fig:raw_error_train_2_aoa}), the estimated AoA (blue) tracks the reference AoA (red) closely across all nine runs. Constant-AoA runs appear as horizontal bands, while varying-AoA runs appear as sloped segments. Deviations are generally smaller at Mach~8 (NZ-1 through NZ-5) than at Mach~5 (NZ-6 through NZ-8).

Figure~\ref{fig:raw_error_test_2_aoa} shows the corresponding results on the withheld test intervals. For constant-AoA cases (NZ-4, NZ-5, NZ-9), the CNN tracks the reference AoA accurately with only minor deviations. For varying-AoA runs, the network continues to follow the overall AoA trend on the withheld interval, but the instantaneous estimates exhibit increased high-frequency variance and, in some cases, residual bias.  Despite never training on data where the vehicle angle is actively decreasing from $6^{\circ}$ to $3^{\circ}$, the network maintains good agreement with the reference AoA at Mach 8 (see NZ-1, NZ-2, NZ-3) and decent agreement at Mach 5 (see NZ-6, NZ-7, NZ-8).

\begin{figure}[h!]
    \centering
    \includegraphics[width=0.70\linewidth]{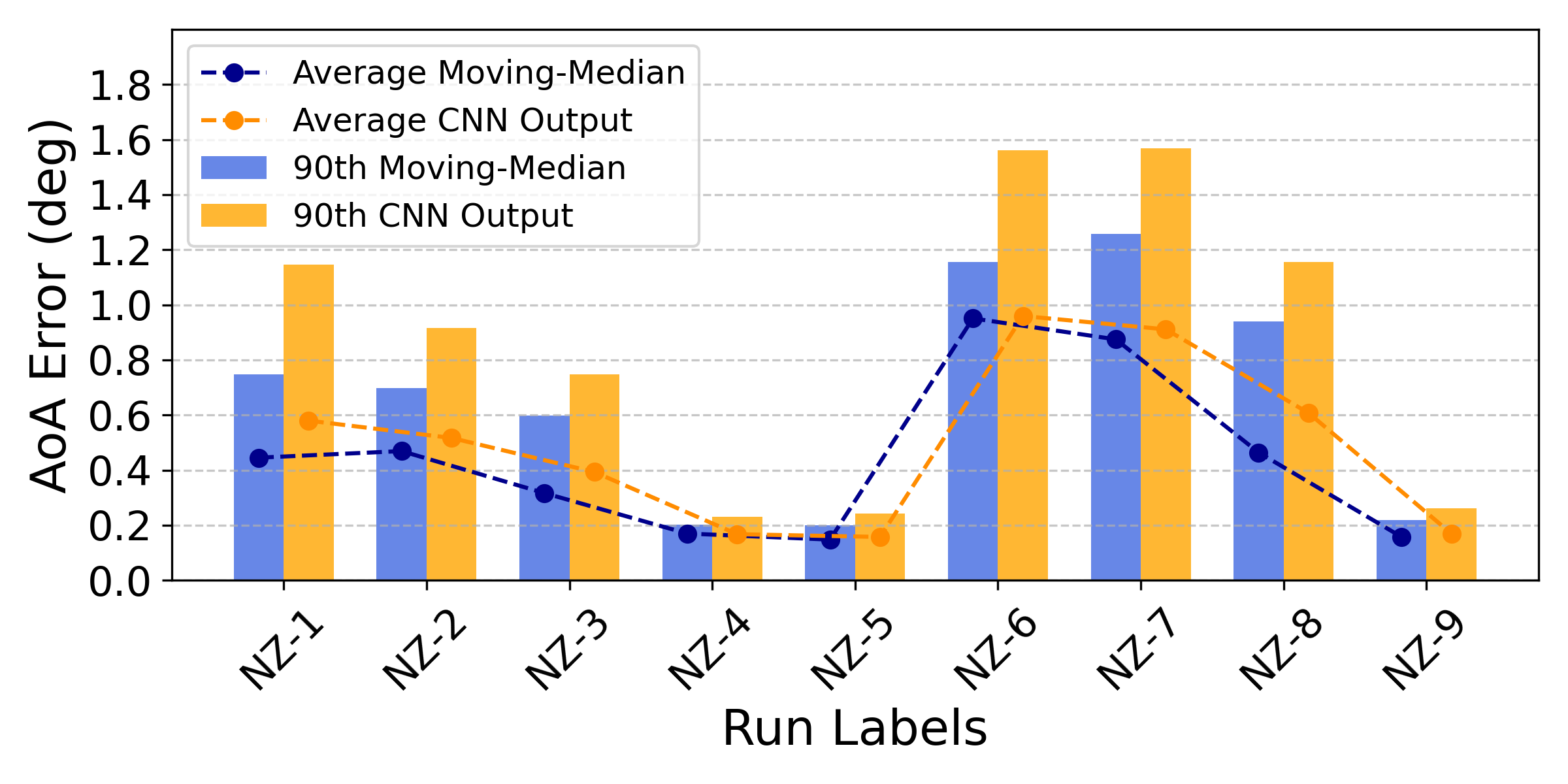}
        \caption{Raw CNN output (orange) and moving-median filtered (blue) AoA estimation errors by run, showing mean (dashed-lines) and 90th percentile (bars). Moving-median results in decreases in 90th percentile errors across all runs, but most significantly for continuously-varying tunnel conditions (Z-1, Z-2, Z-3, Z-6, Z-7, Z-8).}
        \label{fig:compare_median_nonzero_aoa}

\end{figure}

\begin{figure}[h!]
    \centering
    \includegraphics[width=0.70\linewidth]{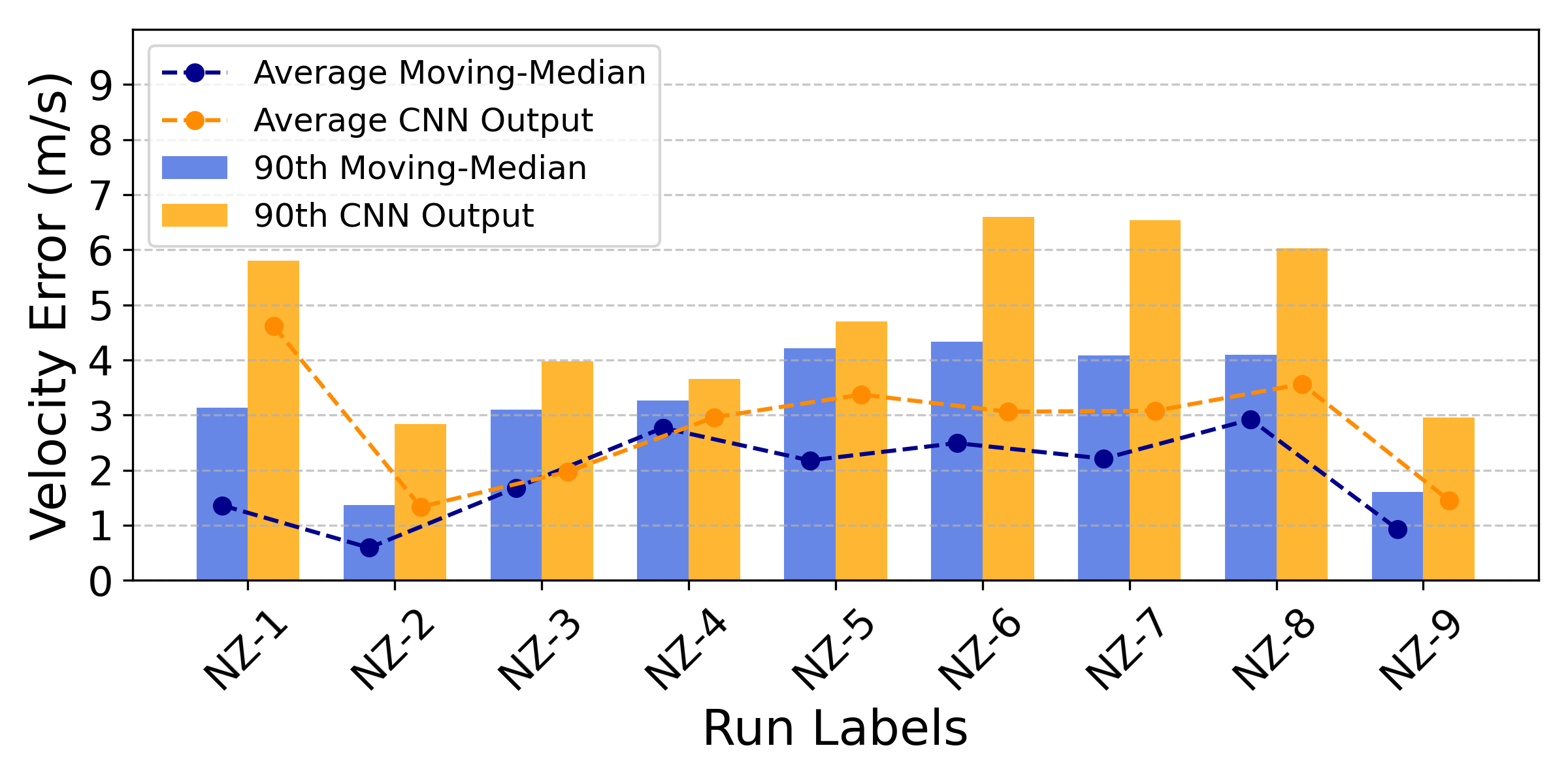}
        \caption{Raw CNN output (orange) and moving-median filtered (blue) velocity magnitude for nonzero AoA estimation errors by run, showing mean (dashed-lines) and 90th percentile (bars). Moving-median results in uniform decreases in 90th percentile.}
        \label{fig:compare_median_nonzero_mag}

\end{figure}

 This observation motivates the sliding median filter introduced in Section~\ref{sec:zeroaoa_free} as a practical and computationally efficient smoothing strategy. Applied to the test data with a 6-sample window, the filtered estimates (shown in orange in Figure~\ref{fig:raw_error_test_2_aoa}) reduce the variance while preserving accuracy. The smoothed trace follows the reference AoA more closely, resulting in estimation improvements across all runs. This is not only a visual observation, but observed acutely from Figure~\ref{fig:compare_median_nonzero_aoa}, which presents a direct comparison of average and 90th percentile performance between the moving-median and the raw CNN outputs. Finally, the sliding-median demonstrates small improvements in velocity estimation across all runs as shown in Figure~\ref{fig:compare_median_nonzero_mag}.

\subsubsection{Differential data requirements for steady vs transient conditions for nonzero AoA flow}\label{sec:nonzeroaoa_free_kfold}

We next repeated the \emph{center-block holdout} study of Section~\ref{sec:zeroaoa_free_kfold} using the nine nonzero AoA runs, again varying holdout test data from 10\% to 50\% in 10\% increments. 
The results suggest that AoA estimation accuracy depends on the degree of overlap between AoA conditions represented in the training segments and those present in the withheld test interval. As is consistent with all presented studies, the training data comprise two disjoint blocks taken from the beginning and end of each run, while a contiguous, temporally centered block is withheld for testing.

\begin{figure}[h!]
    \centering
    \includegraphics[width=1.0\linewidth]{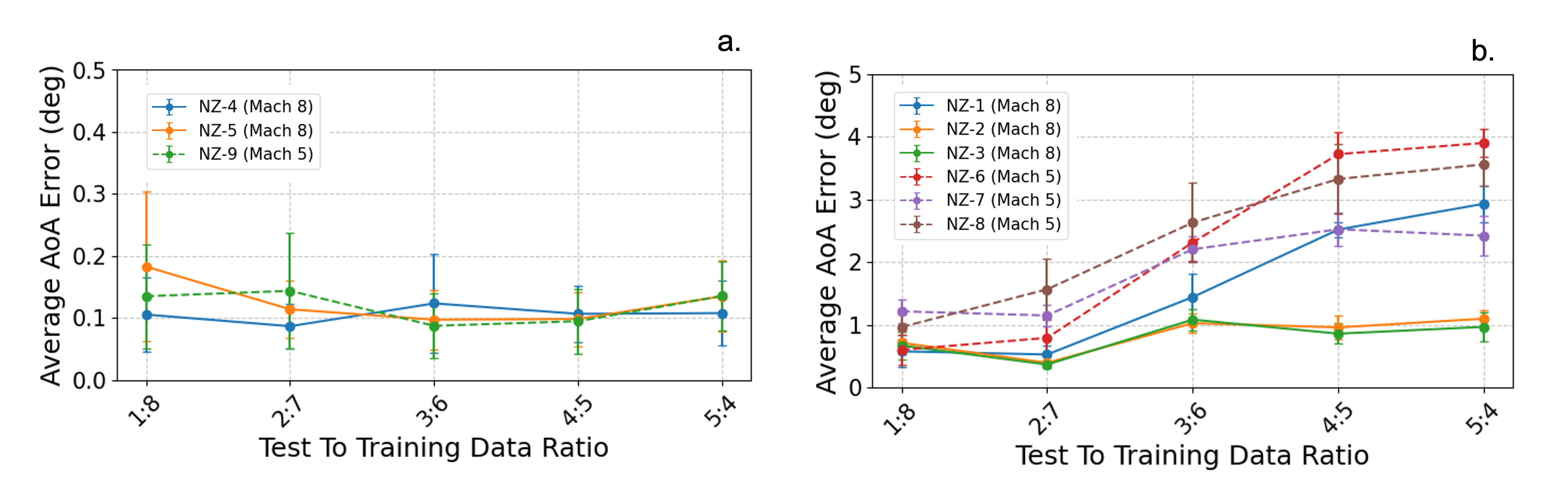}
 \caption{Center-block holdout study for nonzero AoA velocity estimation: a temporally centered contiguous block (10--50\%) is withheld from each run and used for testing. Points show mean absolute error over the withheld block; error bars denote $\pm 1\sigma$ across five retrainings.}
    \label{fig:kfold_nonzeroaoa}
\end{figure}

For constant-condition runs (Fig.~\ref{fig:kfold_nonzeroaoa}a), AoA is fixed, so the response statistics in the withheld block closely match those in the two outer training blocks. Consequently, increasing the withheld fraction from 10\% to 50\% does not produce a systematic increase in AoA error.

For varying-AoA runs (Fig.~\ref{fig:kfold_nonzeroaoa}b), the withheld block corresponds to a distinct portion of the AoA trajectory. As the withheld fraction grows, progressively more mid-trajectory (intermediate-AoA) data are removed from training, reducing overlap between the AoA range covered by the remaining training blocks and the AoA range present in the withheld center block. The resulting increase in error indicates that the CNN can interpolate to unseen time intervals, but its accuracy degrades when the withheld interval contains AoA conditions that are less well represented in the remaining training data.

\section{Results Summary} \label{sec:summary}

Table~\ref{tbl:summary_results} and Table~\ref{tbl:summary_results_rel} consolidate performance across all zero AoA and nonzero AoA test cases (excluding the incremental holdout studies) and compare raw CNN outputs to moving-median post-processing. Errors are reported as absolute error (Table~\ref{tbl:summary_results}) and as relative error \(e_r = |v_{ref}-v_{est}|/|v_{ref}|\) (Table~\ref{tbl:summary_results_rel}). Because the velocity labels are derived from facility estimates with a stated \(\sim 7\%\) systematic uncertainty, the reported velocity errors should be interpreted as agreement with the low-pass filtered \emph{reference velocity} within this experimental campaign rather than absolute freestream velocity accuracy.

Overall, the raw CNN predictions are accurate on average and exhibit the largest degradation in the tail of the error distribution. Applying a short-window moving-median filter yields consistent improvements, with the most pronounced gains occurring in the 90th-percentile and maximum-error statistics. For example, after post-processing the 90th-percentile errors are \(5.97~m/s\) for zero AoA velocity, \(3.70~m/s\) for velocity magnitude under nonzero AoA, and \(1.04^{\circ}\) for AoA. The strong reduction in maximum errors indicates that the moving-median primarily suppresses occasional high-variance/outlier-like estimates while preserving the underlying tracking performance. Together, these results show that the CNN-plus-filter workflow produces stable velocity and AoA estimates over the tested conditions.
\begin{table}[ht]
 \centering
\caption{Summary of Estimation Error Using Test Dataset}
\begin{tabular}{|c|c|c|c|c|c|c|}
    \hline
    & \multicolumn{3}{|c|}{\textbf{CNN Output }} & \multicolumn{3}{|c|}{\textbf{Moving-Median}} \\ 
    \hline
    & \parbox{1.6cm}{\centering \vspace{0.2cm} 0-AoA Vel. \\ $(m/s)$ \vspace{0.2cm}} & 
    \parbox{1cm}{\centering \vspace{0.2cm} AoA \\ $(deg)$ \vspace{0.2cm}} & 
    \parbox{2.5cm}{\centering \vspace{0.2cm} Vel. Magnitude \\ $(m/s)$ \vspace{0.2cm}} & 
    \parbox{1.6cm}{\centering \vspace{0.2cm} 0-AoA Vel. \\ $(m/s)$ \vspace{0.2cm}} & 
    \parbox{1cm}{\centering \vspace{0.2cm} AoA \\ $(deg)$ \vspace{0.2cm}} & 
    \parbox{2.5cm}{\centering \vspace{0.2cm} Vel. Magnitude \\ $(m/s)$ \vspace{0.2cm}} \\
    \hline
    Mean            & 2.68 & 0.50   & 2.82  & 2.27  & 0.44  & 1.90 \\
    90th Percentile & 7.15 & 1.13  & 4.98  & 5.97  & 1.04  & 3.70 \\
    Max             & 13.6 & 4.00  & 93.9  & 8.51  & 1.59  & 7.25 \\
    \hline
\end{tabular}
\label{tbl:summary_results}
\end{table}

\begin{table}[ht]
 \centering
\caption{Summary of Relative Estimation Error Using Test Dataset}
\begin{tabular}{|c|c|c|c|c|c|c|}
    \hline
    & \multicolumn{3}{|c|}{\textbf{CNN Output }} & \multicolumn{3}{|c|}{\textbf{Moving-Median}} \\ 
    \hline
    & \parbox{1.6cm}{\centering \vspace{0.2cm} 0-AoA Vel. \\ (\%) \vspace{0.2cm}} & 
    \parbox{1cm}{\centering \vspace{0.2cm} AoA \\ (\%) \vspace{0.2cm}} & 
    \parbox{2.5cm}{\centering \vspace{0.2cm} Vel. Magnitude \\ (\%) \vspace{0.2cm}} & 
    \parbox{1.6cm}{\centering \vspace{0.2cm} 0-AoA Vel. \\ (\%) \vspace{0.2cm}} & 
    \parbox{1cm}{\centering \vspace{0.2cm} AoA \\ (\%) \vspace{0.2cm}} & 
    \parbox{2.5cm}{\centering \vspace{0.2cm} Vel. Magnitude \\ (\%) \vspace{0.2cm}} \\
    \hline
    Mean            & 0.25 & 9.4  & 0.28  & 0.21 & 8.25 & 0.19 \\
    90th Percentile & 0.67 & 22.2 & 0.51  & 0.54 & 18.8 & 0.38 \\
    Max             & 1.22 & 77.8 & 7.76  & 0.78 & 33.7 & 0.76 \\
    \hline
\end{tabular}
\label{tbl:summary_results_rel}
\end{table}

\section{Conclusion}\label{sec:conclusion}

This work demonstrates proof-of-concept for indirect flow sensing using a dense array of piezoelectric structural-response measurements combined with a convolutional neural network (CNN). Using only interior vibration measurements, the method estimates freestream velocity (relative to the facility-provided low-pass filtered reference velocity) and AoA in controlled wind-tunnel experiments, providing an alternative pathway for aerodynamic state estimation without direct flow instrumentation.

Across the full test dataset, the CNN produces accurate predictions on withheld time intervals, and a short-window moving-median post-processing step improves robustness by suppressing variance and occasional outlier estimates. With the CNN-plus-filter pipeline, 90th-percentile errors are \(5.97~m/s\) under zero AoA, \(3.70~m/s\) for velocity magnitude under nonzero AoA, and \(1.04^{\circ}\) for AoA, yielding stable time histories across both steady-state and time-varying tunnel operations.

A consistent trend across both the zero AoA and nonzero AoA studies is that performance is strongest under constant tunnel conditions and degrades when velocity and(or) AoA varies continuously. The center-block holdout studies clarify this behavior: when a temporally centered test interval is withheld and the model is trained on the remaining early and late portions of the same run, estimation accuracy remains largely unchanged for constant-condition runs (where the withheld interval is statistically similar to the training data) but deteriorates for time-varying runs as the withheld interval becomes less representative of the remaining training data. These results indicate that the CNN can interpolate to unseen time segments, but accuracy depends on having training data that sufficiently cover the conditions present in the withheld interval.

The present study is limited to 16 wind tunnel runs, each corresponding to a distinct set of tunnel conditions. While the results demonstrate generalization within this controlled dataset, it remains an open question whether comparable performance will hold across a broader range of tunnel states and operating conditions. Further study is warranted to determine the extent to which the learned mapping is robust to nuisance parameters (e.g., Reynolds number, pressure/temperature state, density, and other run-to-run environmental differences) versus partially leveraging run-specific signatures correlated with velocity and AoA.

Future work should expand the experimental envelope and explicitly test cross-condition generalization (e.g., training on a subset of runs and evaluating on held-out tunnel conditions), while investigating training strategies that encourage invariance to nuisance parameters. Promising directions include physics-based data augmentation, domain randomization over tunnel-state variables, and incorporating auxiliary measurements (when available) to help disentangle environmental effects from the target quantities.

\section*{Funding Sources}

This work was supported by the Laboratory Directed Research and Development program at
Sandia National Laboratories, a multimission laboratory managed and operated by National
Technology and Engineering Solutions of Sandia LLC, a wholly owned subsidiary of Honeywell
International Inc. for the U.S. Department of Energy’s National Nuclear Security Administration
under contract DE-NA0003525.

\section*{Acknowledgments}
The authors gratefully acknowledge the  Laboratory Directed Research and Development program at
Sandia National Laboratories for its financial support.
We thank Katya Casper and the wind tunnel team for conducting the wind tunnel experiments, Samuel Oxandale for suggesting neural networks for performing velocity estimation, and Jeff Steinfeldt for his help installing the piezoelectrics.

\bibliography{refs}

\begin{thebibliography}{16}
\newcommand{\enquote}[1]{``#1''}
\providecommand{\natexlab}[1]{#1}
\providecommand{\url}[1]{\texttt{#1}}
\providecommand{\urlprefix}{URL }
\expandafter\ifx\csname urlstyle\endcsname\relax
  \providecommand{\doi}[1]{\discretionary{}{}{}https://doi.org/#1}\else
  \providecommand{\doi}[1]{\discretionary{}{}{}\urlstyle{rm}\url{https://doi.org/#1}}\fi

\bibitem[{Corcos(1964)}]{Corcos1963}
Corcos, G.~M., \enquote{The structure of the turbulent pressure field in
  boundary-layer flows,} \emph{Journal of Fluid Mechanics}, Vol.~18, No.~3,
  1964, pp. 353--378.

\bibitem[{Willmarth(1975)}]{Willmarth1975}
Willmarth, W.~W., \enquote{Pressure fluctuations beneath turbulent boundary
  layers,} \emph{Annual Review of Fluid Mechanics}, Vol.~7, No. Volume 7, 1975,
  1975, pp. 13--36.

\bibitem[{Bull(1996)}]{Bull1996}
Bull, M.~K., \enquote{Wall-pressure fluctuations beneath turbulent boundary
  layers: some reflections on forty years of research,} \emph{Journal of Sound
  and Vibration}, Vol. 190, No.~3, 1996, pp. 299--315.

\bibitem[{Chase(1980)}]{Chase1980}
Chase, D., \enquote{Modeling the wavevector-frequency spectrum of turbulent
  boundary layer wall pressure,} \emph{Journal of Sound and Vibration},
  Vol.~70, No.~1, 1980, pp. 29--67.

\bibitem[{Bernardini and Pirozzoli(2011)}]{Bernardini2011}
Bernardini, M., and Pirozzoli, S., \enquote{Wall pressure fluctuations beneath
  supersonic turbulent boundary layers,} \emph{Physics of Fluids}, Vol.~23,
  No.~8, 2011, p. 085102.

\bibitem[{Smith et~al.(2016)Smith, DeChant, Casper, Mesh, and
  R.~J.~Field}]{Smith2016}
Smith, J.~A., DeChant, L.~J., Casper, K.~M., Mesh, M., and R.~J.~Field, J.,
  \enquote{Comparison of a turbulent boundary layer pressure fluctuation model
  to hypersonic cone measurements,} \emph{AIAA Aviation 2016 Forum},
  Washington, D.C., 2016, p. 4047.

\bibitem[{Casper et~al.(2019)Casper, Beresh, Henfling, Spillers, Hunter, and
  Spitzer}]{Casper2019}
Casper, K.~M., Beresh, S.~J., Henfling, J.~F., Spillers, R.~W., Hunter, P., and
  Spitzer, S., \enquote{Hypersonic fluid–structure interactions due to
  intermittent turbulent spots on a slender cone,} \emph{AIAA Journal},
  Vol.~57, No.~2, 2019, pp. 749--759.

\bibitem[{Evans et~al.(2004)Evans, Blotter, and Stephens}]{Evans2004}
Evans, R.~P., Blotter, J.~D., and Stephens, A.~G., \enquote{Flow rate
  measurements using flow-induced pipe vibration,} \emph{Journal of Fluids
  Engineering}, Vol. 126, No.~2, 2004, pp. 280--285.

\bibitem[{Kim and Kim(1996)}]{Kim1996}
Kim, Y., and Kim, Y., \enquote{A three accelerometer method for the measurement
  of flow rate in pipe,} \emph{The Journal of the Acoustical Society of
  America}, Vol. 100, No.~2, 1996, pp. 717--726.

\bibitem[{Thompson et~al.(2010)Thompson, Maynes, and Blotter}]{Thompson2010}
Thompson, A.~S., Maynes, D., and Blotter, J.~D., \enquote{Internal turbulent
  flow induced pipe vibrations with and without baffle plates,} Vol. ASME 2010
  7th International Symposium on Fluid-Structure Interactions, Flow-Sound
  Interactions, and Flow-Induced Vibration and Noise: Volume 3, Parts A and B,
  2010, pp. 649--659.

\bibitem[{Jin et~al.(2020)Jin, Chung, and Park}]{Kim2020}
Jin, J., Chung, Y., and Park, J., \enquote{Development of a flowmeter using
  vibration interaction between gauge plate and external flow analyzed by
  LSTM,} \emph{Sensors}, Vol.~20, No.~20, 2020, p. 5922.

\bibitem[{Nicolas et~al.(2016)Nicolas, Sullivan, and Richards}]{Nicolas2016}
Nicolas, M.~J., Sullivan, R.~W., and Richards, W.~L., \enquote{Large scale
  applications using FBG sensors: determination of in-flight loads and shape of
  a composite aircraft wing,} \emph{Aerospace}, Vol.~3, No.~3, 2016.

\bibitem[{Wada and Tamayama(2019)}]{Wada2019}
Wada, D., and Tamayama, M., \enquote{Wing load and angle of attack
  identification by integrating optical fiber sensing and neural network
  approach in wind tunnel test,} \emph{Applied Sciences}, Vol.~9, No.~7, 2019.

\bibitem[{Kwon et~al.(2019)Kwon, Park, Kim, and Kim}]{Kwon2019}
Kwon, H., Park, Y., Kim, J.-H., and Kim, C.-G., \enquote{Embedded fiber Bragg
  grating sensor--based wing load monitoring system for composite aircraft,}
  \emph{Structural Health Monitoring}, Vol.~18, No.~4, 2019, pp. 1337--1351.

\bibitem[{Vantassel et~al.(2022)Vantassel, Kumar, and Cox}]{Vantassel2022}
Vantassel, J.~P., Kumar, K., and Cox, B.~R., \enquote{Using convolutional
  neural networks to develop starting models for near-surface 2-D full waveform
  inversion,} \emph{Geophysical Journal International}, Vol. 231, No.~1, 2022,
  p. 72–90.

\bibitem[{Jia and Li(2023)}]{Jia2023}
Jia, J., and Li, Y., \enquote{Deep learning for structural health monitoring:
  data, algorithms, applications, challenges, and trends,} \emph{Sensors},
  Vol.~23, No.~21, 2023.

\end{thebibliography}

\end{document}